\documentclass[nohyperref]{article}

\usepackage{microtype}
\usepackage{subcaption}
\usepackage{graphicx}
\usepackage{booktabs} % for professional tables
\usepackage{natbib}

\usepackage{xurl}
\usepackage{hyperref}
\hypersetup{colorlinks=true,breaklinks=true}

\usepackage[accepted]{icml2022}

\usepackage{amsmath}
\usepackage{amssymb}
\usepackage{mathtools}
\usepackage{amsthm}

\usepackage[capitalize,noabbrev]{cleveref}

\theoremstyle{plain}

\theoremstyle{definition}

\theoremstyle{remark}

\usepackage[textsize=tiny]{todonotes}

\usepackage{multirow}
\usepackage{arydshln}
\icmltitlerunning{Context-Aware Drift Detection}

\begin{document}

\twocolumn[
\icmltitle{Context-Aware Drift Detection}




\begin{icmlauthorlist}
\icmlauthor{Oliver Cobb}{yyy}
\icmlauthor{Arnaud Van Looveren}{yyy}
\end{icmlauthorlist}

\icmlaffiliation{yyy}{Seldon Technologies}

\icmlcorrespondingauthor{Oliver Cobb}{oc@seldon.io}

\icmlkeywords{Machine Learning, ICML}

\vskip 0.3in
]



\printAffiliationsAndNotice{}  % leave blank if no need to mention equal contribution

%%% Allow ourselves to toggle between plans for sections and our current version
\begin{abstract}

    When monitoring machine learning systems, two-sample tests of homogeneity form the foundation upon which existing approaches to drift detection build. They are used to test for evidence that the distribution underlying recent deployment data differs from that underlying the historical reference data. Often, however, various factors such as time-induced correlation mean that batches of recent deployment data are not expected to form an i.i.d.\ sample from the historical data distribution. Instead we may wish to test for differences in the distributions conditional on \textit{context} that is permitted to change. To facilitate this we borrow machinery from the causal inference domain to develop a more general drift detection framework built upon a foundation of two-sample tests for conditional distributional treatment effects. We recommend a particular instantiation of the framework based on maximum conditional mean discrepancies. We then provide an empirical study demonstrating its effectiveness for various drift detection problems of practical interest, such as detecting drift in the distributions underlying subpopulations of data in a manner that is insensitive to their respective prevalences. The study additionally demonstrates applicability to ImageNet-scale vision problems.
    
\end{abstract}
\section{Introduction}\label{sec:intro}

Machine learning models are designed to operate on unseen data sharing the same underlying distribution as a set of historical training data. When the distribution changes, the data is said to have drifted and models can fail catastrophically \cite{recht2019imagenet, engstrom2019spatial, hendrycks2019benchmarking, taori2020measuring, barbu2019objectnet}. It is therefore important to have systems in place that detect when drift occurs and raise alarms accordingly \cite{breck2019data, klaise2020monitoring, paleyes2020challenges}.

\begin{figure}
    \centering
    \vspace{-5.75mm}
    \includegraphics[width=0.98\columnwidth, trim={15mm 40mm 31mm 45mm}, clip]{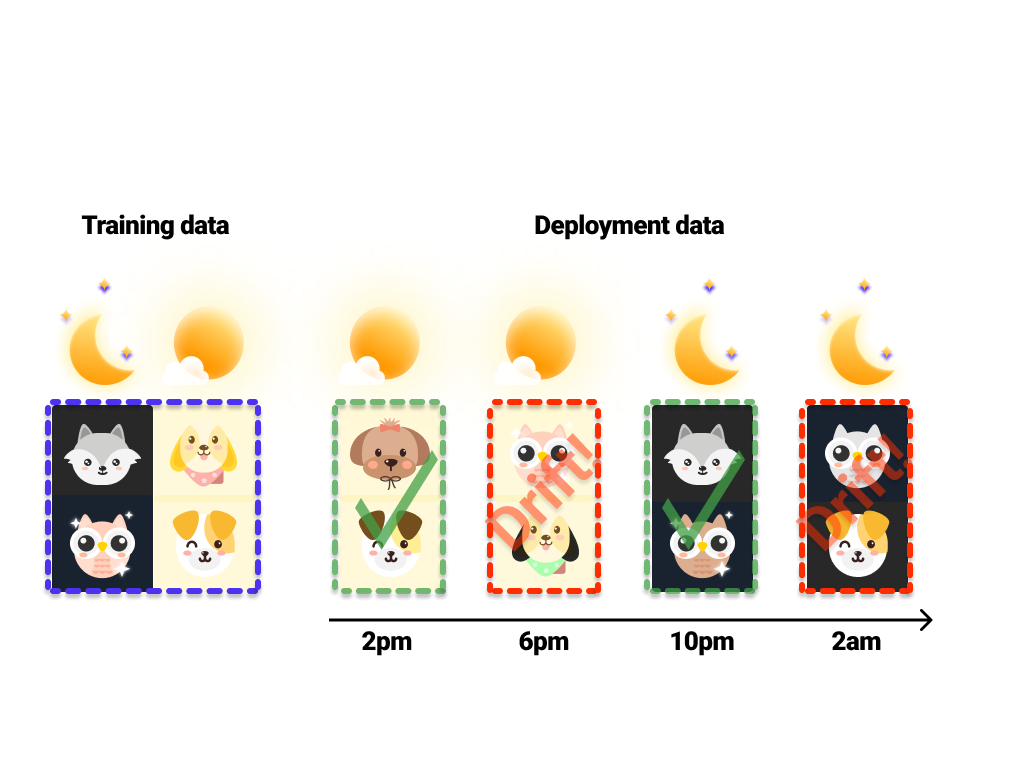}
    \vspace{-0.5mm}
    \caption{Batches of the most recent deployment data may not cover all contexts (e.g. day and night) covered by the training data. We often do not wish for this partial coverage to cause drift detections, but instead to detect other changes not explained by the change/narrowing of context. A nighttime deployment batch should be permitted to contain only wolves and owls, but a daytime deployment batch should not contain any owls.}
    \label{fig:day-night}
    \vspace{-3.5mm}
\end{figure}

Given a model of interest, drift can be categorised based on whether the change is in the distribution of features, the distribution of labels, or the relationship between the two. Approaches to drift detection are therefore diverse, with some methods focusing on one or more of these categories. They invariably, however, share an underlying structure \cite{lu2018learning}. Available data is repeatedly arranged into a set of reference samples, a set of deployment\footnote{Throughout this paper we use ``deployment'' data to mean the unseen data on which a model was designed to operate.} samples, and a test of equality is performed under the assumption that the samples within each set are i.i.d.\ Methods then vary in the notion of equality chosen to be repeatedly tested, which may be defined in terms of specific moments, model-dependent transformations, or in the most general distributional sense. \cite{gretton2008dataset, gretton2012mmd, tasche2017fisher, lipton2018detecting, page1954cusum, gama2004learning, baena2006early, bifet2007learning, wang2015concept, quionero2009dataset, rabanser2019failing,cobb2021sequential}. Each test evaluates a test statistic capturing the extent to which the two sets of samples deviate from the chosen notion of equality and make a detection if a threshold is exceeded. Threshold values are set such that, under the assumptions of i.i.d.\ samples and strict equality, the rate of false positives is controlled. However, in practice windows of deployment data may stray from these assumptions in ways deemed to be permissible.

As a simple example, illustrated in Figure~\ref{fig:day-night}, consider a computer vision model operating on sequentially arriving images. The distribution underlying the images is known to change depending on the time of day, throughout which lighting conditions change. The distribution underlying a nighttime batch of deployment samples differs from that underlying the full training set, which also contains daytime images. Despite this, the model owner does not wish to be inundated with alarms throughout the night reminding them of this fact. In this situation the time of day (or lighting condition) forms important context which the practitioner is happy to let vary between windows of data. They are interested only in detecting deviations from equality that can not be attributed to a change in this context. Although this simple example may be addressed by comparing only to relevant subsets of the reference data, more generally changes in context are distributional and such simple approaches can not be effectively applied.

Deviations from the i.i.d.\ assumption are the norm rather than the exception in practice. An application may be used by different age groups at different times of the week. Search engines experience surges in similar queries in response to trending news stories. Food delivery services expect the distribution of orders to differ depending on the weather. In all of these cases there exists important context that existing approaches to drift detection are not equipped to account for. A common response is to decrease the sensitivity of detectors so that such context changes cause fewer unwanted detections. This, however, hampers the detector's ability to detect the potentially costly changes that \textit{are} of interest. With this in mind, our contributions are to: 
\begin{enumerate}
    \item Develop a framework for drift detection that allows practitioners to specify context permitted to change between windows of deployment data and only detects drift that can not be attributed to such context changes.
    \item Recommend an effective and intuitive instantiation of the framework based on the maximum mean discrepancy (MMD) and explore connections to popular MMD-based two-sample tests.
    \item Explain and demonstrate the applicability of the framework to various drift detection problems of practical interest.
    \item Make an implementation available to use as part of the open-source Python library \texttt{alibi-detect} \cite{Van_Looveren_Alibi_Detect_Algorithms_2022}.
\end{enumerate}
% (i)~Develop a framework for drift detection that allows practitioners to specify context permitted to change between windows of deployment data and only detects drift that can not be attributed to such context changes. (ii) Recommend an effective and intuitive instantiation of the framework based on the maximum mean discrepancy (MMD) and explore connections to popular MMD-based two-sample tests. (iii)  Explain and demonstrate the applicability of the framework to various drift detection problems of practical interest. (iv) Make an implementation available to use as part of the open-source Python library \texttt{alibi-detect} \cite{Van_Looveren_Alibi_Detect_Algorithms_2022}.

% \begin{itemize}\compresslist
%     \item Develop a framework for drift detection that allows practitioners to specify context permitted to change between windows of deployment data and only detects drift that can not be attributed to such context changes.
%     \item Recommend an effective and intuitively simple instantiation of the framework based on maximum mean discrepancy (MMD) and explore connections to popular MMD-based two-sample tests.
%     \item Explain and demonstrate the applicability of the framework to various drift detection problems of practical interest.
% \end{itemize}
\section{Background and Notation}

We briefly review two-sample tests for homogeneity and treatment effects. To make clear the connections between the two we adopt treatment-effect notation for both. We focus on methods applicable to multivariate distributions and sensitive to differences or effects in the general distributional sense, not only in certain moments or projections.

\subsection{Two-sample tests of homogeneity}\label{sec:homo}

Let $(\Omega, \mathcal{F}, P)$ denote a probability space with associated random variables $X: \Omega \rightarrow \mathcal{X}$ and $Z: \Omega \rightarrow \{0,1\}$. We subscript $P$ to denote distributions of random variables such that, for example, $P_X$ denotes the distribution of $X$. Consider an i.i.d.~sample $(\mathbf{x}, \mathbf{z})=\{(x_i,z_i)\}_{i=1}^n$ from $P_{X,Z}$ and the two associated samples $\mathbf{x}^0=\{x_i^0\}_{i=1}^{n_0}$ and $\mathbf{x}^1=\{x_i^1\}_{i=1}^{n_1}$ from $P_{X_0} := P_{X|Z=0}$ and $P_{X_1} := P_{X|Z=1}$ respectively. A two-sample test of homogeneity is a statistical test of the null hypothesis $h_0: P_{X_0}=P_{X_1}$ against the alternative $h_1: P_{X_0} \neq P_{X_1}$. The test starts by specifying a test statistic $\hat{t}:\mathcal{X}^{n_0} \times \mathcal{X}^{n_1} \rightarrow \mathbb{R}$, typically an estimator of a distance $D(P_{X_0}, P_{X_1})$, expected to be large under $h_1$ and small under $h_0$. To test at significance level (false positive probability) $\alpha$, the observed value of the test statistic is computed along with the probability $p=P_{h_0}(T>\hat{t}(\mathbf{x}^0,\mathbf{x}^1))$ that such a large value of the test statistic would be observed under $h_0$. If $p<\alpha$ then $h_0$ is rejected.

Although effective alternatives exist \cite{lopezpaz2017revisiting, ramdas2017wasserstein, bu2018lsdd}, we focus on kernel-based test statistics \cite{harchaoui2007testing, gretton2008dataset, gretton2009fast, fromont2012kernels, gretton2012mmd, gretton2012optimal, chwialkowski2015fast, jitkrittum2016interpretable, liu2020learning} which are particularly popular due to their applicability to any domain $\mathcal{X}$ upon which a kernel $k: \mathcal{X} \times \mathcal{X} \rightarrow \mathbb{R}$, capturing a meaningful notion of similarity, can be defined. The most common example is to let $\hat{t}$ be an estimator of the squared MMD $D(P_{X_0}, P_{X_1})=||\mu_{P_0}-\mu_{P_1}||^2_{\mathcal{H}_k}$ \cite{gretton2012mmd}, which is the distance between the distributions' kernel mean embeddings \cite{muandet2017kernel} in the reproducing kernel Hilbert space $\mathcal{H}_k$. The squared MMD admits a consistent (although biased) estimator of the form 
\begin{equation}
\begin{split}\label{eqn:mmd}
    \hat{t}(\mathbf{x}^0, \mathbf{x}^1) &= \frac{1}{n_0^2}\sum_{i,j} k(x^0_i,x^0_j) + \frac{1}{n_1^2}\sum_{i,j} k(x^1_i,x^1_j) \\
     &-\frac{2}{n_0n_1}\sum_{i,j} k(x^0_i,x^1_j).
\end{split}
\end{equation}
In cases such as this, where the distribution of the test statistic $\hat{t}$ under the null distribution $h_0$ is unknown, it is common to use a permutation test to obtain an accurate estimate $\hat{p}$ of the unknown p-value $p$. This compares the observed value of the test statistic $\hat{t}$ against a large number of alternatives that could, with equal probability under the null hypothesis, have been observed under random reassignments of the indexes $\{z_i\}_{i=1}^n$ to instances $\{x_i\}_{i=1}^n$. 
\subsection{Two-sample tests for treatment effects}\label{sec:te}

A related problem is that of inferring treatment effects. Here, instead of asking whether $X$ is independent of $Z$ we ask whether it is causally affected by $Z$. To illustrate, now consider $Z$ a treatment assignment and $X$ an outcome of interest. We may write $X=X^0(1-Z) + X^1Z$, 
where both the observed outcome and the counterfactual outcome corresponding to the alternative treatment assignment are considered. If, as is common in observational studies, the treatment assignment $Z$ is somehow guided by $X$, then the distribution of $X$ will depend on $Z$ even if the treatment is ineffective. The dependence will be non-causal however. In such cases, to determine the causal effect of $Z$ on $X$ it is important to control for covariates $C$ through which $Z$ might depend on $X$. Supposing we can identify such covariates, i.e. a random variable $C: \Omega \rightarrow \mathcal{C}$ satisfying the following condition of \textit{unconfoundedness},
\begin{equation}\tag{A1}\label{assumption:1}
    Z \perp\!\!\!\perp (X^0,X^1) \;|\; C,
\end{equation}
then differences between the distributions of $X^0$ and $X^1$ can be identified from observational data. This is because unconfoundedness ensures $P_{X_0|C}:=P_{X|C,Z=0}=P_{X^z|C,Z=0}=P_{X^0|C}$, and likewise for $Z=1$. Henceforth we assume the unconfoundedness assumption holds and use $P_{X_z|C}$ to refer to both $P_{X|C,Z=z}$ and $P_{X^z|C}$.

A common summary of effect size is the average treatment effect $\text{ATE}=E[X^1]-E[X^0]$  \cite{rosenbaum1983central}, which is the expectation of the conditional average treatment effect (CATE) $U(c)=E[X^1|C=c]-E[X^0|C=c]$ with respect to the marginal distribution $P_C$. More generally however we may be interested in effects beyond the mean and consider a treatment effect to exist if the conditional distributions $P_{X_0|C=c}(\cdot)$ and $P_{X_1|C=c}(\cdot)$ are not equal almost everywhere (a.e.)\ with respect to $P_C$ \cite{lee2009nonparametric, chang2015nonparametric, shen2019estimation}. In this more general setting the effect size can be summarised by defining a conditional \textit{distributional} treatment effect (CoDiTE) function $U_D(c)=D(P_{X_0|C=c}, P_{X_1|C=c})$ and marginalising over $P_C$ to obtain what we refer to as an average distributional treatment effect (ADiTE) $E[D(P_{X_0|C}, P_{X_1|C})]$. This is the expected distance, as measured by $D$, between two $C$-measurable random variables. For either the ATE or ADiTE quantities to be well defined requires a second assumption of \textit{overlap},
\begin{equation}\label{assumption:2}\tag{A2}
    0 < e(c) := P(Z=1|C=c) < 1 \;\;P_{C}\text{-a.e.},
\end{equation}
to avoid the inclusion of quantities conditioned on events of zero probability. The function $e:\mathcal{C}\rightarrow [0,1]$ is often referred to as the propensity score.

Although various CoDiTE functions have been proposed \cite{hohberg2020treatment, chernozhukov2013inference, briseno2020flexible}, to the best of our knowledge only that of \citet{park2021conditional} straightforwardly (through a kernel formulation) generalises to multivariate
% \footnote{Only the $\mathcal{X}\subseteq\mathbb{R}$ case is considered in \citet{park2021conditional}, but this is assumedly due to their specific interests and their introduction of interpretability methods requiring the $\mathcal{X}\subseteq\mathbb{R}$ assumption.} 
and non-euclidean domains for both $\mathcal{X}$ and $\mathcal{C}$. They choose $D$ to be the squared MMD such that the associated CoDiTE
\begin{equation}\label{eqn:mmd-codite}
    U_{\text{MMD}}(c)=||\mu_{X_0|C=c}-\mu_{X_1|C=c}||^2_{\mathcal{H}_k}
\end{equation} 
 is the squared distance in $\mathcal{H}_k$ between the mean embeddings of $P_{X_0|C=c}$ and $P_{X_1|C=c}$, which are each well defined under the overlap assumption \ref{assumption:2}. The associated ADiTE $E[U_{\text{MMD}}(C)]$ is then equal to the expected distance between the \textit{conditional mean embeddings} (CMEs) \mbox{$\mu_{X_0|C}=E[k(X_0,\cdot)|C]$} and $\mu_{X_1|C}=E[k(X_1,\cdot)|C]$, which are $C$-measurable random variables in $\mathcal{H}_k$. Here CMEs are defined using the measure theoretic formulation 
 which \citet{park2020measure} introduce as preferable to the operator-theoretic formulation of \citet{song2009hilbert} for various reasons. \citet{singh2020reproducing} study CoDiTE-like quantities within the operator-theoretic framework.

 To estimate $U_{\text{MMD}}(c)$ \citet{park2021conditional} introduce a covariate kernel \mbox{$\l:\mathcal{C}\times\mathcal{C}\rightarrow\mathbb{R}$} and perform regularised operator-valued kernel regression to obtain the estimator
\begin{align}\label{eqn:mmd-codte-est}
    \hat{U}_{\text{MMD}}(c)&=\mathbf{l}^{\top}_0(c)\mathbf{L}_{\lambda_0}^{-1}\mathbf{K}_{0,0}\mathbf{L}_{\lambda_0}^{-\top}\mathbf{l}^{\top}_0(c) \nonumber \\ 
    &+ \mathbf{l}^{\top}_1(c)\mathbf{L}_{\lambda_1}^{-1}\mathbf{K}_{1,1}\mathbf{L}_{\lambda_1}^{-\top}\mathbf{l}^{\top}_1(c) \\ 
    &- 2\mathbf{l}^{\top}_0(c)\mathbf{L}_{\lambda_0}^{-1}\mathbf{K}_{0,1}\mathbf{L}_{\lambda_1}^{-\top}\mathbf{l}^{\top}_1(c), \nonumber
\end{align}
where $\mathbf{L}^{-1}_{\lambda_z}=(\mathbf{L}_{z, z}+\lambda_z n_z\mathbf{I})^{-1}$ is a regularised inverse of the kernel matrix $\mathbf{L}_{z, z}$ with $(i,j)$-th entry $l(c^z_i,c^z_j)$ and $\mathbf{l}_z(c)$ is the vector with $i$-th entry $l(c^z_i,c)$. This estimator is consistent if $k$ and $l$ are bounded, $l$ is universal and $\lambda_0$ and $\lambda_1$ decay at slower rates than $\mathcal{O}(n_0^{-\frac12})$ and $\mathcal{O}(n_1^{-\frac12})$ respectively. Moreover the associated ADiTE $E[U_{\text{MMD}}(C)]$ can be consistently estimated via the Monte Carlo estimator
\begin{equation}\label{eqn:adite-est}
    \hat{t}(\mathbf{x},\mathbf{c}, \mathbf{z})=\frac1n\sum_{i=1}^n\hat{U}_{\text{MMD}}(c_i).
\end{equation}
\citet{park2020measure} use this estimator to test the null hypothesis that there exists no distributional treatment effect of $Z$ on $X$. Unlike for tests of homogeneity however, the p-value can not straightforwardly be estimated using a permutation test. This is because a value of the unpermuted test statistic that is extreme relative to the permuted variants may result from a dependence of $Z$ on $(X^0,X^1)$ that ceases to exist under permutation of $\{z_i\}_{i=1}^n$. By the unconfoundedness assumption the dependence must pass through $C$ and can therefore be preserved by reassigning treatments as $z'_i\sim\text{Ber}(e(c_i))$, instead of naively permuting the instances. Under the null hypothesis, the distribution of the original statistic and test statistics computed via this treatment reassignment procedure are then equal \cite{rosenbaum1984conditional}, allowing p-values to be estimated in the usual way. Implementing this procedure requires using $\{(c_i,z_i)\}_{i=1}^n$ to fit a classifier $\hat{e}(c)$ approximating the propensity score $e(c)$.

%By the unconfoundedness assumption the dependence must pass through $C$ however and therefore ``permutations'' respecting the dependence can be obtained by reassigning treatments as $z'_i\sim\text{Ber}(e(c_i))$ for $i=1,...,n$. Under the null hypothesis, the distribution of the original and ``permuted'' test statistics are then equal \cite{rosenbaum1984conditional}, allowing p-values to be obtained in the usual way. Implementing this procedure requires using $\{(c_i,z_i)\}_{i=1}^n$ to fit a classifier $\hat{e}(c)$ approximating the propensity score $e(c)$.

% This is because a value of the unpermuted test statistic that is extreme relative to the permuted variants may result from a dependence of $Z$ on $(X_0,X_1)$ that ceases to exist under permutation $\mathbf{z\}$ of $\{z_i\}_{i=1}^n$. By the unconfoundedness assumption the dependence must pass through $C$ however and therefore permutations respecting the dependence can be obtained by reassigning treatments as $z'_i\sim\text{Ber}(e(c_i))$ for $i=1,...,n$. Under the null hypothesis, the distribution of the original and permuted test statistics are then equal \cite{rosenbaum1984conditional}, allowing p-values to be obtained in the usual way. Implementing this procedure requires using $\{(c_i,z_i)\}_{i=1}^n$ to fit a classifier $\hat{e}(c)$ approximating the propensity score $e(c)$.

\section{Context-Aware Drift Detection}

Suppose we wish to monitor a model $M: \mathcal{X} \rightarrow \mathcal{Y}$ mapping features $X\in\mathcal{X}$ onto labels $Y \in\mathcal{Y}$. Existing approaches to drift detection differ in the statistic $S(X,Y;M)$ chosen to be monitored\footnote{The unavailability of deployment labels and fact that any change in the distribution of features could cause  performance degradation, makes $S(X,Y;M)=X$ a common choice.}
for underlying changes in distribution $P_S$. This involves applying tests of homogeneity to samples $\{s^0_i\}_{i=1}^{n_0}$ and $\{s^1_i\}_{i=1}^{n_1}$ as described in Section~\ref{sec:homo}. In this section we will introduce a framework that affords practitioners the ability to augment samples with realisations of an associated context variable $C\in\mathcal{C}$, whose distribution is permitted to differ between reference and deployment stages. It is then the conditional distribution $P_{S|C}$ that is monitored by applying methods adapted from those in Section~\ref{sec:te} to contextualised samples $\{(s^0_i,c^0_i)\}_{i=1}^{n_0}$
and $\{(s^1_i,c^1_i)\}_{i=1}^{n_1}$. Recalling the deployment scenarios discussed in Section~\ref{sec:intro}, the underlying motivation is that the reference set often corresponds to a wider variety of contexts than a specific deployment batch. In other words $P_{C_1}$ may differ from $P_{C_0}$ in a manner such that the support of the former may be a strict subset of that of the latter. In such scenarios, which are common in practice, we postulate that practitioners instead wish for their detectors to satisfy the following desiderata:
\begin{description}
    \item[D1:]\label{D1} The detector should be completely insensitive to changes in the data distribution that \textit{can} be attributed to changes in the distribution of the context variable.
    \item[D2:]\label{D2} The detector should be sensitive to changes in the data distribution that \textit{can not} be attributed to changes in the distribution of the context variable.
    \item[D3:]\label{D3} The detector should prioritise being sensitive to changes in the data distribution in regions that are highly probable under the deployment distribution.
\end{description}
Before describing our framework for drift detection that satisfies the above desiderata, we describe a number of drift detection problems of practical interest, corresponding to different choices of $S(X,Y;M)$ and $C$ for which we envisage our framework being particularly useful.
\begin{enumerate}
    \item Features $X$ conditional on an indexing variable $t$ -- such as time, lighting or weather -- informed by domain specific knowledge.
    \item Features $X$ conditional on the relative prevalences of known subpopulations. This would allow changes to the proportion of instances falling into pre-existing modes of the distribution whilst requiring the distribution of each mode to remain constant.
    \item Features $X$ conditional on model predictions $M(X)$. An increased frequency of certain predictions should correspond to the expected change in the covariate distribution rather than the emergence of a new concept. Similarly, conditioning on a notion of model uncertainty $H(M(X))$ would allow increases in model uncertainty due to covariate drift into familiar regions of high aleatoric uncertainty (often fine) to be distinguished from that into unfamiliar regions of high epistemic uncertainty (often problematic).
    \item Labels $Y$ conditional on features $X$. Although deployment labels are rarely available, this would correspond to explicitly detecting \textit{concept drift} where the underlying change is in the conditional distribution $P_{Y|X}$.
\end{enumerate}

\subsection{Context-Aware Drift Detection with ADiTT Estimators}\label{sec:framework}

Consider a set of contextualised reference samples $\{(s^0_i,c^0_i)\}_{i=1}^{n_0}$ and deployment samples $\{(s^1_i,c^1_i)\}_{i=1}^{n_1}$. Rather than making the assumption that each set forms an i.i.d.~sample from their underlying distributions and testing for equality, we first make the much weaker assumption that $\{(s_i,c_i, z_i)\}_{i=1}^{n}$ constitues an i.i.d.~sample from $P_{S,C,Z}$, where $Z\in\{0,1\}$ is a domain indicator with reference samples corresponding to $Z=0$ and deployment samples to $Z=1$. We then make the stronger assumption that each i.i.d.~sample admits the following generative process
\begin{align}
    \begin{gathered}
    Z \sim P_Z, \\
    C \sim P_{C|Z}, \\
    S^0 \sim P_{S^0|C}, \;\;
    S^1 \sim P_{S^1|C} \\
    S = S^0(1-Z) + S^1Z.
    \end{gathered}
\end{align}
Intuitively we can consider this as relaxing the assumption that $\{s_i^0\}_{i=1}^{n_0}$ and $\{s_i^1\}_{i=1}^{n_1}$ are each i.i.d.~samples to them each being i.i.d.~conditional on their respective contexts $\{c_i^0\}_{i=1}^{n_0}$ and $\{c_i^1\}_{i=1}^{n_1}$. We are then interested in testing for differences between $P_{S^0|C}$ and $P_{S^1|C}$. Note that focusing on these context-conditional distributions allows the marginal distributions $P_{S_0}$ and $P_{S_1}$ underling $\{s_i^0\}_{i=1}^{n_0}$ and $\{s_i^1\}_{i=1}^{n_1}$ to differ, so long as the difference can be attributed to a difference between the distributions $P_{C_0}$ and $P_{C_1}$ underlying $\{c_i^0\}_{i=1}^{n_0}$ and $\{c_i^1\}_{i=1}^{n_1}$. Importantly, the above process satisfies the unconfoundedness condition of $Z \perp\!\!\!\perp (S^0,S^1) \;|\; C$, such that $P_{S_z|C}:=P_{S|C,Z=z}=P_{S^z|C}$. This allows us to apply adapted versions of the methods described in Section~\ref{sec:te} to test the null and alternative hypotheses
\begin{description}
    \item[$h_0:$] $P_{S_0|C=c}(\cdot)=P_{S_1|C=c}(\cdot) \;\;P_{C_1}$-almost everywhere,
    % \item[$h_1:$] $P_{S_0|C=c}(\cdot)\neq P_{S_1|C=c}(\cdot) \;\;P_{C_1}$-almost nowhere.
    \item[$h_1:$] $P_{C_1}\big(\{c\in\mathcal{C}: P_{S_0|C=c}(\cdot)\neq P_{S_1|C=c}(\cdot)\}\big)>0$.
\end{description}
Note here that we interest ourselves only in differences between $P_{S_0|C=c}(\cdot)$ and $P_{S_1|C=c}(\cdot)$ at contexts supported by the deployment context distribution $P_{C_1}$. It would not be possible, or from the practitioner's point of view desirable (D3), to detect differences at contexts that are not possible under the deployment distribution $P_{C_1}$, regardless of their likelihood under the reference distribution $P_{C_0}$.

 To test the above hypotheses first recall that we may use a CoDiTE function $U_D(c)$, as introduced in Section~\ref{sec:te}, to assign contexts $c$ to distances between corresponding conditional distributions $P_{S_0|C=c}$ and $P_{S_1|C=c}$. Secondly note that testing the above hypotheses is equivalent to testing whether, for deployment instances specifically and controlling for context, their status as deployment instances causally affected the distribution underlying their statistics. This focus on deployment instances specifically means that we are not interested in an effect summary such as ADiTE that averages a context-conditional effect summary $U_D(c)$ over the full context distribution $P_{C}$, but instead in an effect summary that averages a context-conditional summary over the deployment context distribution $P_{C_1}$. This narrowing of focus is not uncommon in analogous cases in causal inference where practitioners wish to focus on the average effect of a treatment on those who actually received it -- the average effect \textit{on the treated} -- commonly abbreviated as ATT. 
We therefore refer to the distributional version, $E[U_D(C)|Z=1]$, as the ADiTT associated with $D$.

Assuming $D$ is a probability metric or divergence, (thereby satisfying $D(P,Q)\geq0$ with equality iff $P=Q$), the ADiTT is non-zero if and only if the null hypothesis $h_0$ fails to hold. We may therefore consider it an oracular test statistic and note that any consistent estimator $\hat{t}$ can be used in a consistent test of the hypothesis. In practice however we must estimate it using a finite number of samples and the fact that the estimate is a weighted average w.r.t. $P_{C_1}$ explicitly adheres to desiderata D3. We use \citet{rosenbaum1984conditional}'s conditional resampling scheme, as described in Section~\ref{sec:te}, to estimate the p-value summarising the extremity of $\hat{t}$ under the null. Thankfully when estimating the p-value associated with an estimate of ADiTT (or ATT), we can relax the overlap assumption required when estimating a p-value associated with an estimate of ADiTE (or ATE). We instead must only satisfy the condition of \textit{weak overlap}
\begin{equation}\label{assumption:3}\tag{A3}
    0 < e(c) = P(Z=1|C=c) < 1 \;\; P_{C_1}\text{-a.e.},
\end{equation}
which is a necessary relaxation given we do not expect all contexts supported by the reference distribution to be supported by the deployment distribution. 

Our general framework for context-aware drift detection is described in Algorithm~\ref{alg:cadd}. Implementation, however, requires a suitable choice of CoDiTE function and procedure for estimating the associated ADiTT. CATE \mbox{(i.e. $U_D(c)=E[S_1|C=c]-E[S_0|C=c]$)}, for which associated ATT estimators are well established, can be considered a particularly simple choice. Recall, however, that changes harmful to the performance of a deployed model may not be easily identifiable though the mean and therefore we focus on distributional alternatives.

 \begin{algorithm}[tb]
    \caption{Context-Aware Drift Detection}
    \label{alg:cadd}
 \begin{algorithmic}
    \STATE {\bfseries Input:} Statistics, contexts and domains $(\mathbf{s},\mathbf{c}, \mathbf{z})$, an ADiTT estimator $\hat{t}$, significance level $\alpha$, number of permutations $n_{\text{perm}}$.
    \STATE Compute $\hat{t}$ on $(\mathbf{s},\mathbf{c}, \mathbf{z})$.
    \STATE Use $(\mathbf{c}, \mathbf{z})$ to fit a probabilistic classifier $\hat{e}(c)$ that approximates the propensity score $e(c)$. 
    \FOR{$i=1$ {\bfseries to} $n_{\text{perm}}$}
    \STATE Reassign domains as $\mathbf{z}_i \sim \text{Bernoulli}(\hat{e}(\mathbf{c}))$.
    \STATE Compute $\hat{t}_i$ on $(\mathbf{s},\mathbf{c}, \mathbf{z}_i)$
    \ENDFOR
    \STATE {\bfseries Output:} p-value $\hat{p}= \frac{1}{n_{\text{perm}}}\sum_{i=1}^{n_{\text{perm}}} 1\{\hat{t}_i>\hat{t}\}$
 \end{algorithmic}
 \end{algorithm}

\subsection{MMD-based ADiTT Estimation}\label{sec:mmd-aditt}

\begin{figure*}
    \centering
    \includegraphics[width=0.94\textwidth, trim={-40 0 0 0}, clip]{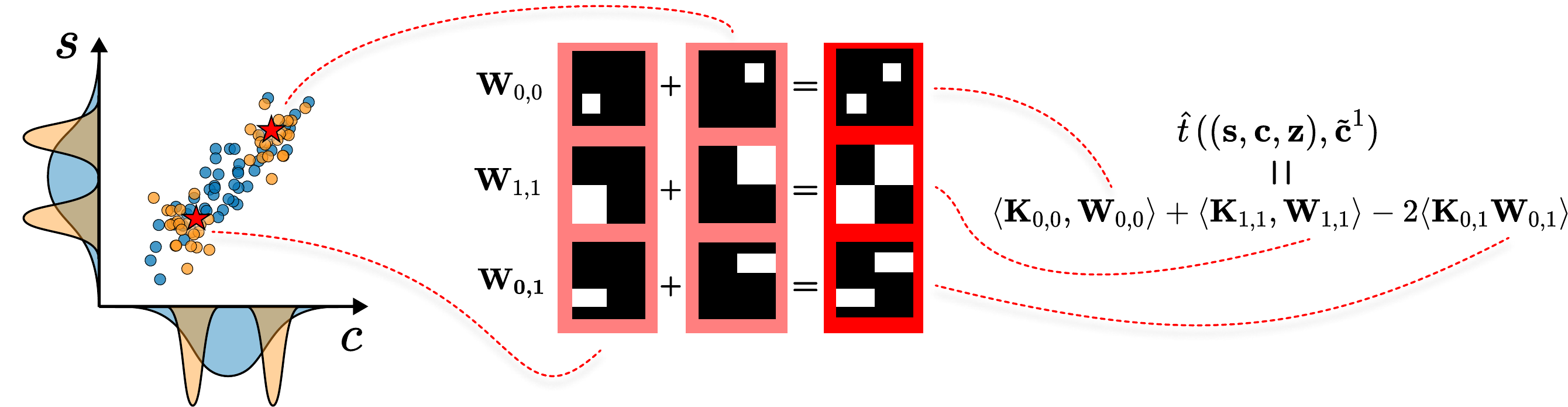}
    \caption{Illustration of the computation of the MMD-based ADiTT statistic, where the difference between distributions underlying the reference (blue) and deployment (orange) data $s$ can be attributed to a change in distribution of context $c$. Computation can be thought of as considering a number of held out deployment samples (red stars), and for each one computing a weighted MMD where only reference and deployment samples with similar contexts significantly contribute. These weighted MMDs are then averaged to form the test statistic. To visualise weight matrices we sort the samples in ascending order w.r.t. $c$ and, for $\textbf{W}_{0,1}$ for example, show the $(j,k)$-th entry as white if the similarity $k(s^0_j,s^1_k)$ is to significantly contribute.}
    \label{fig:pipeline}
\end{figure*}

In this section we recommend \citet{park2021conditional}'s MMD-based CoDiTE function and adapt their corresponding estimator for use within the framework described above. We make this recommendation for several reasons. Firstly, it allows for statistics and contexts residing in any domains $\mathcal{S}$ and $\mathcal{C}$ upon which meaningful kernels can be defined. Secondly, the computation consists primarily of manipulating kernel matrices, which can be implemented efficiently on modern hardware. Thirdly, the procedure is simple and intuitive and closely parallels the MMD-based approach that is widely used for two-sample testing.
% In this section we recommend \citet{park2021conditional}'s MMD-based CoDiTE function and adapt their corresponding estimator for use within the framework described above. We make this recommendation for several reasons:
% \begin{itemize}
% \compresslist
%     \item It allows conditional drift detection for statistics and contexts residing in any domains $\mathcal{S}$ and $\mathcal{C}$ upon which meaningful kernels can be defined.
%     \item The computation consists primarily of manipulating kernel matrices, which can be implemented efficiently on modern hardware.
%     \item The procedure is simple and intuitive and closely parallels the MMD-based approach that is widely used for two-sample testing.
% \end{itemize}

Recall from Section~\ref{sec:te} that \citet{park2021conditional} define, for a given kernel $k:\mathcal{S}\times\mathcal{S}\rightarrow\mathbb{R}$, the MMD-based CoDiTE  $U_{\text{MMD}}(c)=||\mu_{S_0|C=c}-\mu_{S_1|C=c}||^2_{\mathcal{H}_k}$ that captures the squared distance between the kernel mean embeddings of conditional distributions $P_{S_0|C=c}$ and $P_{S_1|C=c}$. 
They show that $U_{\text{MMD}}(c)$ and the associated ADiTE $E[U_{\text{MMD}}(C)]$ can each be consistently estimated by Equations~\ref{eqn:mmd-codte-est} and~\ref{eqn:adite-est} respectively. However, the context-aware drift detection framework instead requires estimation of the ADiTT \mbox{$E[U_{\text{MMD}}(C)|Z=1]$}. We note that this can be achieved with the alternative estimator
\begin{equation}
    \hat{t}(\mathbf{s},\mathbf{c}, \mathbf{z})=\frac{1}{n_1}\sum_{i=1}^{n_1}\hat{U}_{\text{MMD}}(c^1_i),
\end{equation}
which averages only over deployments contexts $\{c_i^1\}_{i=1}^{n_1}$, rather than all contexts $\{c_i\}_{i=1}^{n}$. We further note that although this estimator is consistent, and therefore asymptotically unbiased, averaging over estimates of CoDiTE conditioned on the same contexts used in estimation introduces bias at finite sample sizes. We instead recommend the test statistic
\begin{equation}\label{eqn:aditt-est}
    \hat{t}((\mathbf{s},\mathbf{c}, \mathbf{z}), \tilde{\mathbf{c}}^1)=\frac{1}{n_h}\sum_{i=1}^{n_h}\hat{U}_{\text{MMD}}(\tilde{c}^1_i),
\end{equation}
where a portion $\tilde{\mathbf{c}}^1\in\mathcal{C}^{n_h}$ of the deployment contexts (e.g. 25\%) are held out to be conditioned on whilst the rest are used for estimating the corresponding CoDiTEs. A further motivation for this modification is that conditioning on and averaging over all possible contexts carries a high computational cost that we found unjustified.

\subsubsection{Setting Regularisation Parameters $\lambda_0, \lambda_1$}

In Section~\ref{sec:te} we noted that $\hat{U}_{\text{MMD}}(c)$ is a consistent estimator of $U_{\text{MMD}}(c)$ if $\lambda_0$ and $\lambda_1$ decay at slower rates than $\mathcal{O}(n_0^{-\frac12})$ and $\mathcal{O}(n_1^{-\frac12})$ respectively. In practice however sample sizes are fixed and values for $\lambda_0$ and $\lambda_1$ must be chosen. These parameters arise as regularisation parameters in an operator-valued kernel regression, where functions $\{k(s^z_i,\cdot)\}_{i=1}^{n_z}$ are regressed against contexts $\{c^z_i\}_{i=1}^{n_z}$ to obtain an estimator of $\mu_{S_z|C}=E[k(S^z,\cdot)|C]$. We propose using $k$-fold cross-validation to identify regularisation parameters that minimise the validation error in this regression problem. Full details can be found in Appendix~\ref{add:lam_tune}.

\subsubsection{Relationship to MMD Two-Sample Tests}

To facilitate illustrative comparisons to traditional MMD-based tests of homogeneity, we first note that Equation~\ref{eqn:mmd} can be rewritten in matrix form as
\begin{equation}\label{eqn:mmd-langle}
    t(\mathbf{s}^0, \mathbf{s}^1)=\langle \mathbf{K}_{0,0}, \mathbf{W}_{0,0} \rangle + \langle \mathbf{K}_{1,1}, \mathbf{W}_{1,1} \rangle - 2\langle \mathbf{K}_{0,1} \mathbf{W}_{0,1} \rangle,
\end{equation}
 where, for $u,v\in\{0,1\}$, $\mathbf{K}_{u,v}$ denotes the kernel matrix with $(j,k)$-th entry $k(s^u_j,s^v_k)$ and $\mathbf{W}_{u,v}$ is a uniform weight matrix with all entries equal to $(n_un_v)^{-1}$. We now additionally note that we can rewrite Equation~\ref{eqn:aditt-est} in exactly the same form but with $\textbf{W}_{u,v}=\sum_{i=1}^{n_h} \textbf{W}_{u,v,i}$, where
\begin{equation}
  \textbf{W}_{u,v,i} =  (\mathbf{L}_{\lambda_u}^{-1}\mathbf{l}_u(\tilde{c}^1_i))(\mathbf{L}_{\lambda_v}^{-1}\mathbf{l}_v(\tilde{c}^1_i))^{\top}.
\end{equation}
Here $\textbf{W}_{u,v,i}$ can be viewed as an outer product between $\mathbf{l}_u(\tilde{c}^1_i)$ and $\mathbf{l}_v(\tilde{c}^1_i)$, assigning weight to pairs $(c^u_j,c^v_k)$ that are both similar to $\tilde{c}_i^1$, but adjusted via $\mathbf{L}_{\lambda_u}^{-1}$ and $\mathbf{L}_{\lambda_v}^{-1}$ such that the weight is less if $c^u_j$ (resp. $c^v_k$) has many similar instances in $\mathbf{c}^u$ (resp. $\mathbf{c}^v$). This adjustment is important to ensure that, for example, the combined weight assigned to comparing $s^0_j$ to $s^1_k$, both with contexts similar to $\tilde{c}_i^1$, does not depend on the number of reference or deployment instances with similar contexts. If there are fewer such instances they receive higher weight to compensate. This has the effect of controlling for $P_{C_0}$ and $P_{C_1}$ in the CoDiTE estimates. The dependence on $P_{C_1}$ returns when we average over the held out contexts $\tilde{\mathbf{c}}^1$. We visualise this process for estimating the MMD-based ADiTT in Figure~\ref{fig:pipeline}. For illustrative purposes we consider only two held out contexts, visualise the corresponding weight matrices $\textbf{W}_{u,v,i}$ for $u,v\in\{0,1\}$ and $i\in\{1,2\}$ and show how they are summed to form the matrices $\textbf{W}_{u,v}$ used by Equation~\ref{eqn:mmd-langle}. In particular note the intuitive block structure showing that only similarities $k(s^u_j,s^v_k)$ between instances with similar contexts are weighted and contribute to the test statistic $\hat{t}$. By contrast the weight matrices used in an estimate of MMD used by two-sampling testing approaches would be fully white.

% \begin{algorithm}[tb]
%     \caption{Context Aware Drift Detection}
%     \label{alg:cadd}
%  \begin{algorithmic}
%     \STATE {\bfseries Input:} Data, contexts and domains $(\mathbf{x},\mathbf{c}, \mathbf{z})$, held out deployment contexts $\tilde{\mathbf{c}}^1$, kernel $k$ on $\mathcal{X}$, kernel $l$ on $\mathcal{C}$, significance level $\alpha$, number of permutations $n_{\text{perm}}$, regularisation parameters $(\lambda_0, \lambda_1)$.
%     \STATE Compute $\hat{t}$ on $(\mathbf{x},\mathbf{c}, \mathbf{z})$ as in Equation~\ref{eqn:aditt-est}.
%     \STATE Use $(\mathbf{c}, \mathbf{z})$ to fit a probabilistic classifier $\hat{e}(c)$ that approximates the propensity score $e(c)$. 
%     \FOR{$i=1$ {\bfseries to} $n_{\text{perm}}$}
%     \STATE Reassign domains as $\mathbf{z}_i \sim \text{Bernoulli}(\hat{e}(\mathbf{c}))$.
%     \STATE Compute $\hat{t}_i$ on $(\mathbf{x},\mathbf{c}, \mathbf{z}_i)$
%     \ENDFOR
%     \STATE {\bfseries Output:} p-value $\hat{p}= \frac{1}{n_{\text{perm}}}\sum_{i=1}^{n_{\text{perm}}} 1\{\hat{t}_i>\hat{t}\}$
%  \end{algorithmic}
%  \end{algorithm}
\section{Experiments}\label{sec:exps}
This section will show that using the MMD-based ADiTT estimator of Section~\ref{sec:mmd-aditt} within the framework developed in Section~\ref{sec:framework} results in a detector satisfying desiderata D1-D3. This will involve showing that the resulting detector is calibrated when there has been no change in the distribution $P_S$ \textit{and} when there is a change which can be attributed to a change in the context distribution $P_C$. This prevents comparisons to conventional drift detectors that are not designed to detect drift for the latter case. We therefore first develop a baseline that generalises the principle underlying ad-hoc approaches that might be considered by practitioners faced with changing context. 

% We then compare to this baseline first on two simple examples that facilitate visual as well as quantitative analysis, before considering an ImageNet problem demonstrating that the approach scales to real-world computer vision problems of practical interest.
% We therefore first develop and present a baseline that we believe is similar in spirit to ad-hoc approaches practitioners might consider using when faced with changing context. We then compare to this baseline first on two simple examples that facilitate visual as well as quantitative analysis, before considering an ImageNet problem demonstrating that the approach scales to real-world computer vision problems of practical interest.

Suppose a batch of deployment instances $\mathbf{s}^1$ has contexts $\mathbf{c}^1$ (such as time) contained within a strict subset $[\min(\mathbf{c}^1),\max(\mathbf{c}^1)] \subset [\min(\mathbf{c}^0),\max(\mathbf{c}^0)]$ of those covered by the reference distribution (e.g. 1 of 24 hours). Practitioners might here perform a two-sample test of the deployment batch against the subsample of reference instances with contexts contained within the same interval. More generally the practitioner wishes to perform the test with a subset of the reference data that is sampled such that its underlying context distribution matches $P_{C_1}$. Knowledge of $P_{C_0}$ and $P_{C_1}$ would allow the use of rejection sampling to obtain such a subsample and subsequent application of a two-sample test would provide a perfectly calibrated detector for our setting. We therefore consider, as a baseline we refer to as MMD-Sub, rejection sampling using density estimators $\hat{P}_{C_0}$ and $\hat{P}_{C_1}$. We cannot fit the density estimators using the samples being rejection sampled. We therefore use the held-out portion of samples that the MMD-based ADiTT method uses to condition on. For the closest possible comparison we fit kernel density estimators using the same kernel as the ADiTT method and apply the MMD-based two-sample test  described in Section~\ref{sec:homo}. Further details on this baseline can be found in Appendix~\ref{app:baseline}.

Evaluating detectors by performing multiple runs and reporting false positive rates (FPRs) and true positive rates (TPRs) at fixed significance levels results in high-variance performance measures that vary depending on the levels chosen. We instead evaluate power using AUC; the area under the receiver operating characteristic curve (ROC) of FPR plotted against TPR across all significance levels. More powerful detectors obtain higher AUCs. To evaluate calibration we similarly capture the FPR across all significance levels using the Kolmogorov-Smirnov (KS) distance between the set of obtained p-values and $\text{U}[0,1]$: the distribution of p-values for a perfectly calibrated detector. We contextualise KS distances in plots by shading the interval (0.046,  0.146): the 95\% confidence interval of the KS distance computed using p-values actually sampled from $U[0,1]$.  We use plots to present key trends contained within results and defer full tables, as well as more detailed descriptions of experimental procedures, to Appendices \ref{app:aditt}-\ref{sec:app_expperiments}. This includes, in Appendix~\ref{app:adite}, a discussion of ablations performed to confirm the importance of using the adapted estimator of ADiTT, rather than \citet{park2021conditional}'s estimator of ADiTE. For kernels we use Gaussian RBFs with bandwidths set using the median heuristic \cite{gretton2012mmd}. 
For MMD-ADiTT we fit $\hat{e}(c)$ using kernel logistic regression.

\subsection{Controlling for domain specific context}\label{sec:4_1}
\begin{figure}
\begin{subfigure}{0.49\linewidth}
      \includegraphics[trim={3mm 6mm 12mm 0mm}, clip, width=1.0\linewidth]{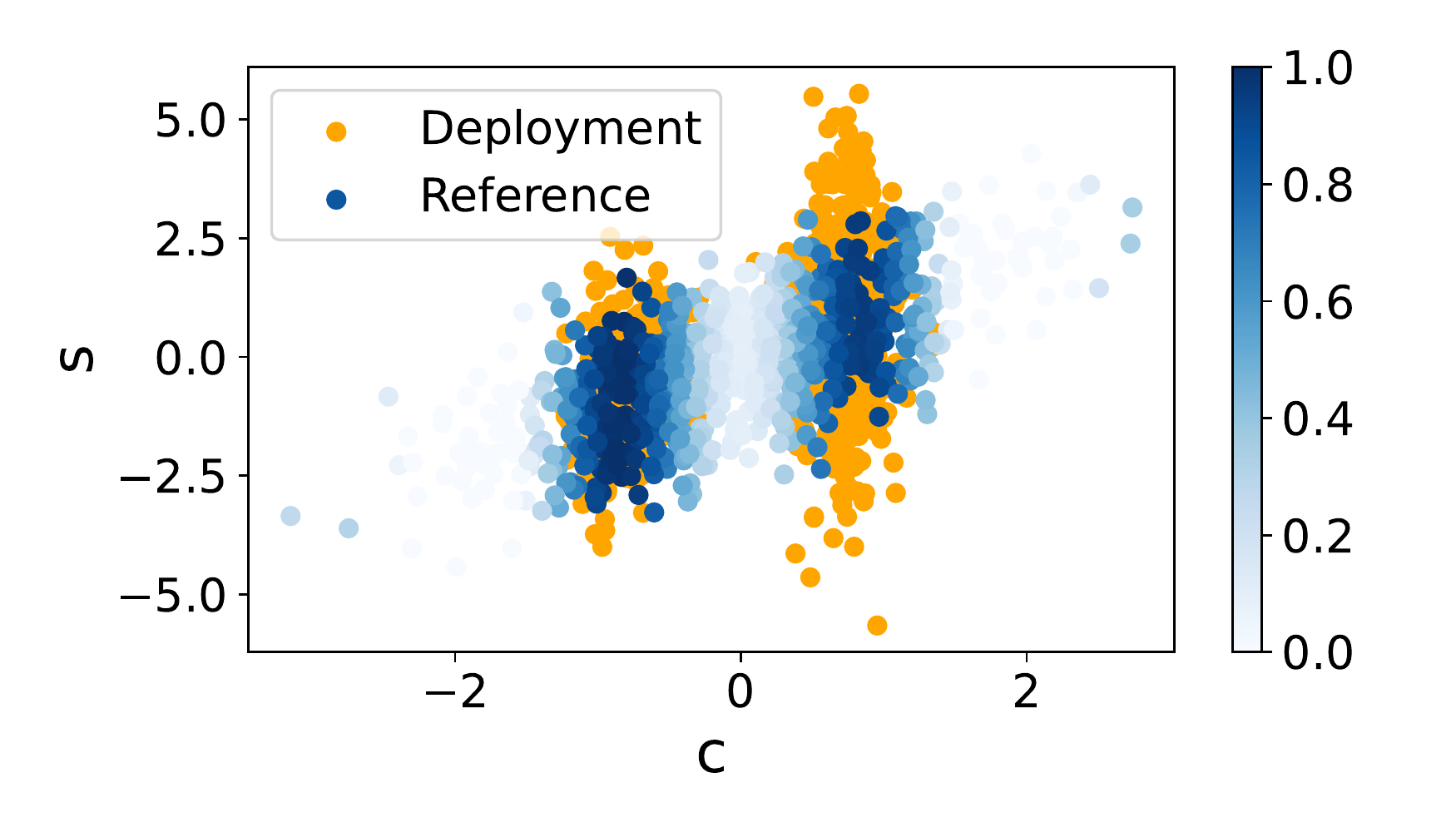}
    %   \caption{\small{.}}
    \end{subfigure}
    \begin{subfigure}{0.49\linewidth}
      \centering
      \includegraphics[trim={3mm 6mm 12mm 0mm}, clip, width=1.0\linewidth]{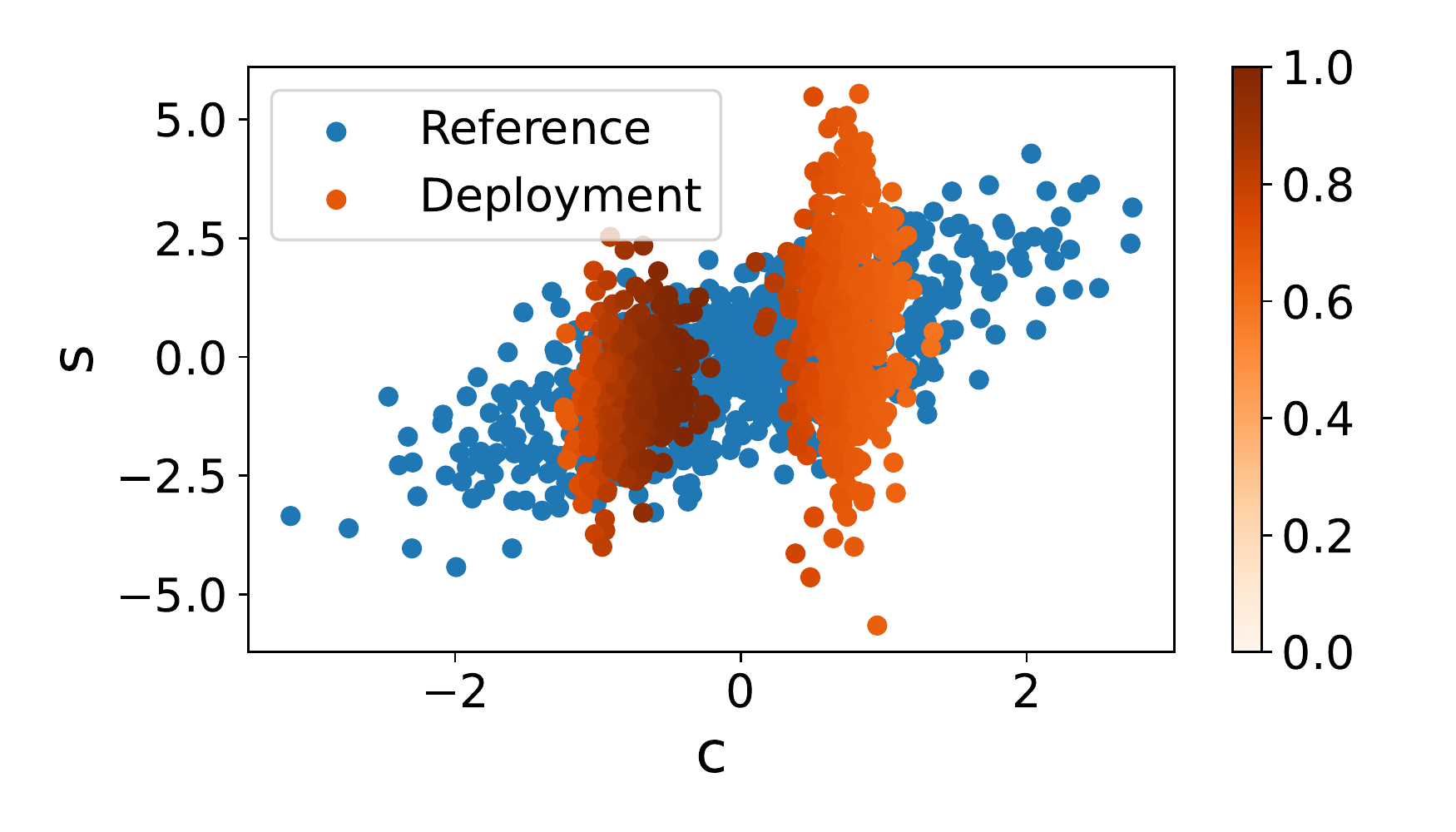}
    %   \caption{\small{.}}
    \end{subfigure}
    \caption{Visualisation of the weight attributed by MMD-ADiTT to comparing each reference sample to the set of deployment samples (left) and vice versa (right). Only reference samples with contexts in the support of the deployment contexts significantly contribute. Weight here refers to the corresponding row/column sum of $W_{0,1}$.}
    \label{fig:tv_marginal_weights}
    % \vspace{-1mm}
\end{figure}

\begin{figure}
    \centering
    \includegraphics[trim={0 6mm 0 0mm}, clip, width=\linewidth]{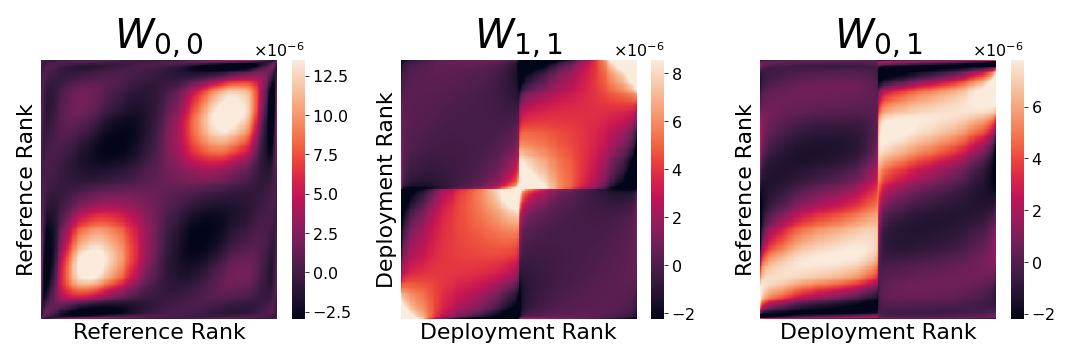}
    \caption{Visualisation of the weight matrices used in computation of the MMD-ADiTT when deployment contexts fall into two disjoint modes. The block-like structure that emerges when ordering the samples by context confirms that only similarities between instances with contexts in the same deployment mode contribute.}
    \label{fig:tv_heatmaps}
    \vspace{-2mm}
\end{figure}

\begin{figure}
    \begin{subfigure}{0.49\linewidth}
        \centering
        \includegraphics[trim={5mm 5mm 3mm 0mm}, clip, width=1.0\linewidth]{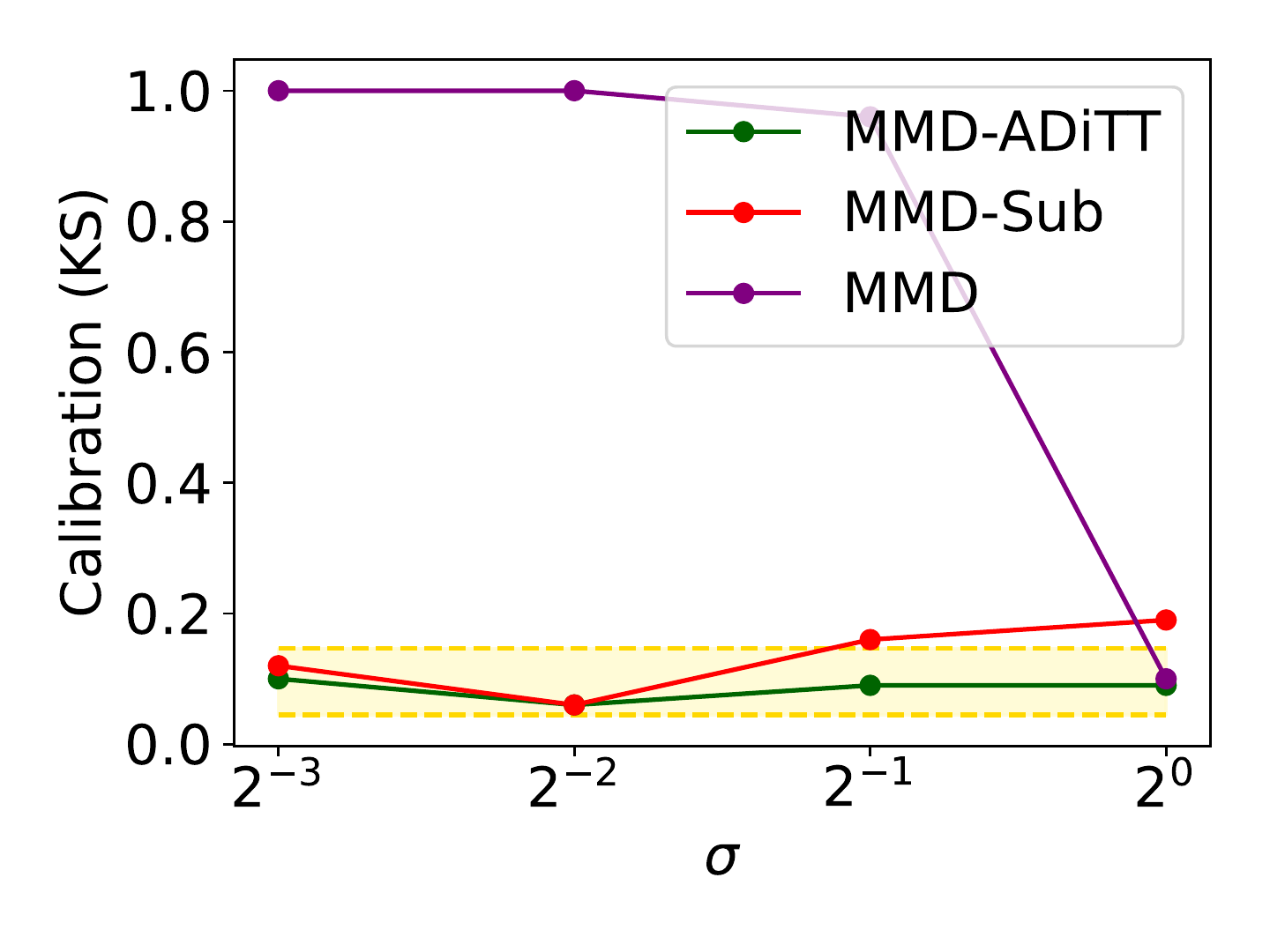}
        % \caption{}
        \label{fig:tv_calib_nulls_null}
      \end{subfigure}
    \begin{subfigure}{0.49\linewidth}
        \centering
        \includegraphics[trim={5mm 5mm 3mm 0mm}, clip, width=1.0\linewidth]{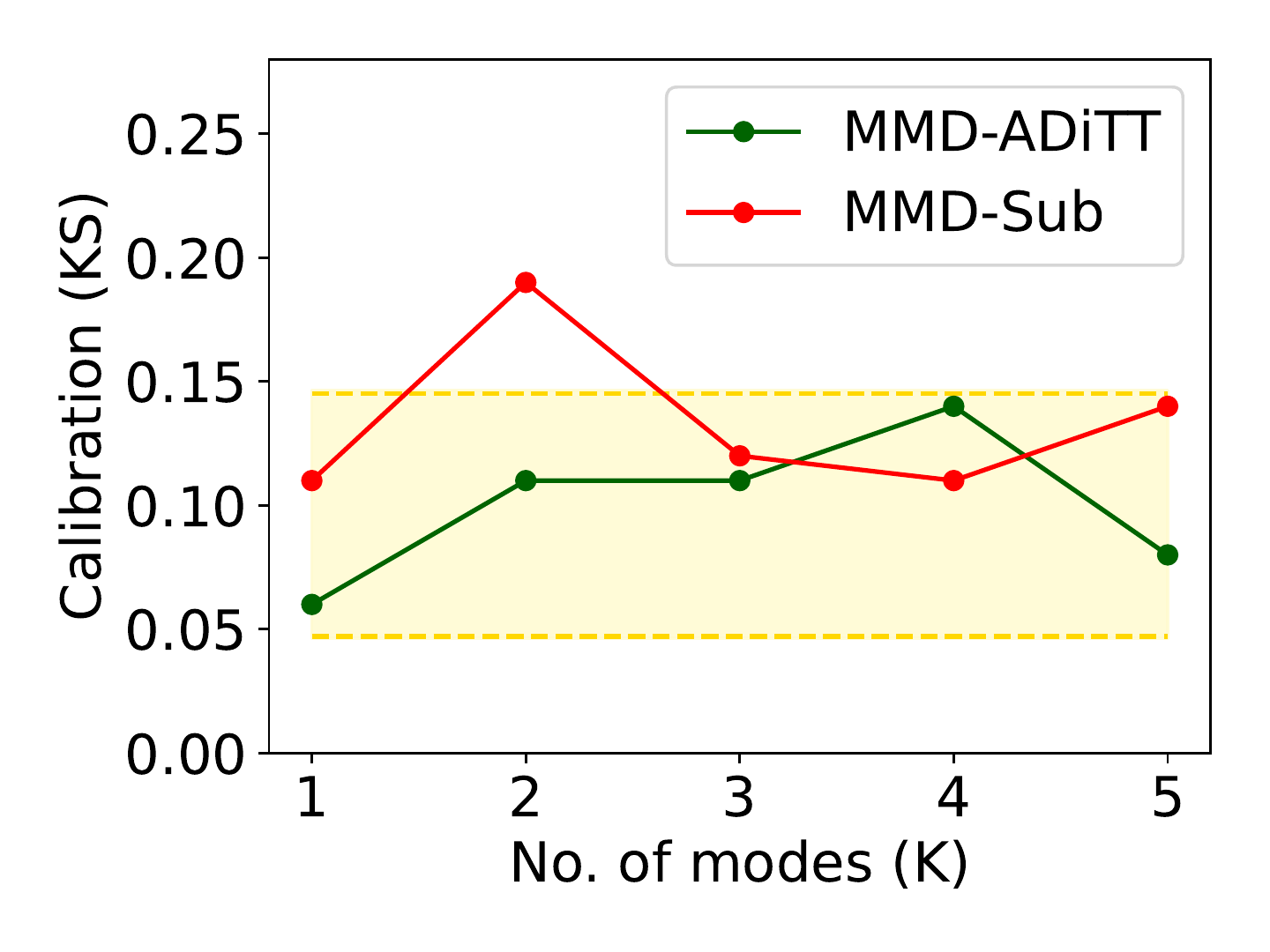}
        % \caption{\small{.}}
        \label{fig:tv_calib_nulls}
      \end{subfigure}
    %   \caption{Caption}
      \vspace{-2.5mm}
      \caption{Plots of the calibration of detectors as (left) the context distribution $P_{C_1}$ gradually narrows from $N(0,1)$ to $N(0,\sigma^2)$ and (right) completely changes to a mixture of Gaussians with K modes.}
      \label{fig:tv_calib}
    \end{figure}

This example is designed to correspond to problems where we wish to allow changes in domain specific context, such as time or weather. To facilitate visualisations we consider univariate statistics $S\in \mathbb{R}$ and contexts $C\in\mathbb{R}$. For the reference distribution we take $S_0 \sim N(C,1)$ and $C_0 \sim N(0,1)$. For changes in the context distributions we consider for $P_{C_1}$ both a simple narrowing from $N(0,1)$ to $N(0,\sigma^2)$ where $\sigma<1$ and a more complex change to a mixture of Gaussians with $K$ modes. Figure~\ref{fig:tv_calib} shows that MMD-ADiTT remains perfectly calibrated across all settings and MMD-Sub is also strongly calibrated. Unsurprisingly, even a slight narrowing of context causes conventional two-sample tests to become wildly uncalibrated in our setting.
% First, to make explicit the motivation for context aware detectors whilst demonstrating the suitability of the baseline, we keep the same context-conditional distribution $S_1 \sim N(C,1)$ and consider the alternative context distributions $C_1 \sim N(0,\sigma^2)$ for $\sigma \in \{0.2,0.4,0.6,0.8,1.0\}$. Figure~\ref{fig:tv_calib} confirms that all three approaches are well calibrated for $\sigma=1$, where the marginal reference and deployment statistic distributions are equal at $S_0 \sim N(0,2)$, but only MMD-ADiTT and MMD-Sub remain well calibrated as $\sigma$ decreases.

\begin{figure}
    \begin{center}
    \includegraphics[width=0.95\linewidth, trim={2mm 2mm 2mm 1mm}, clip]{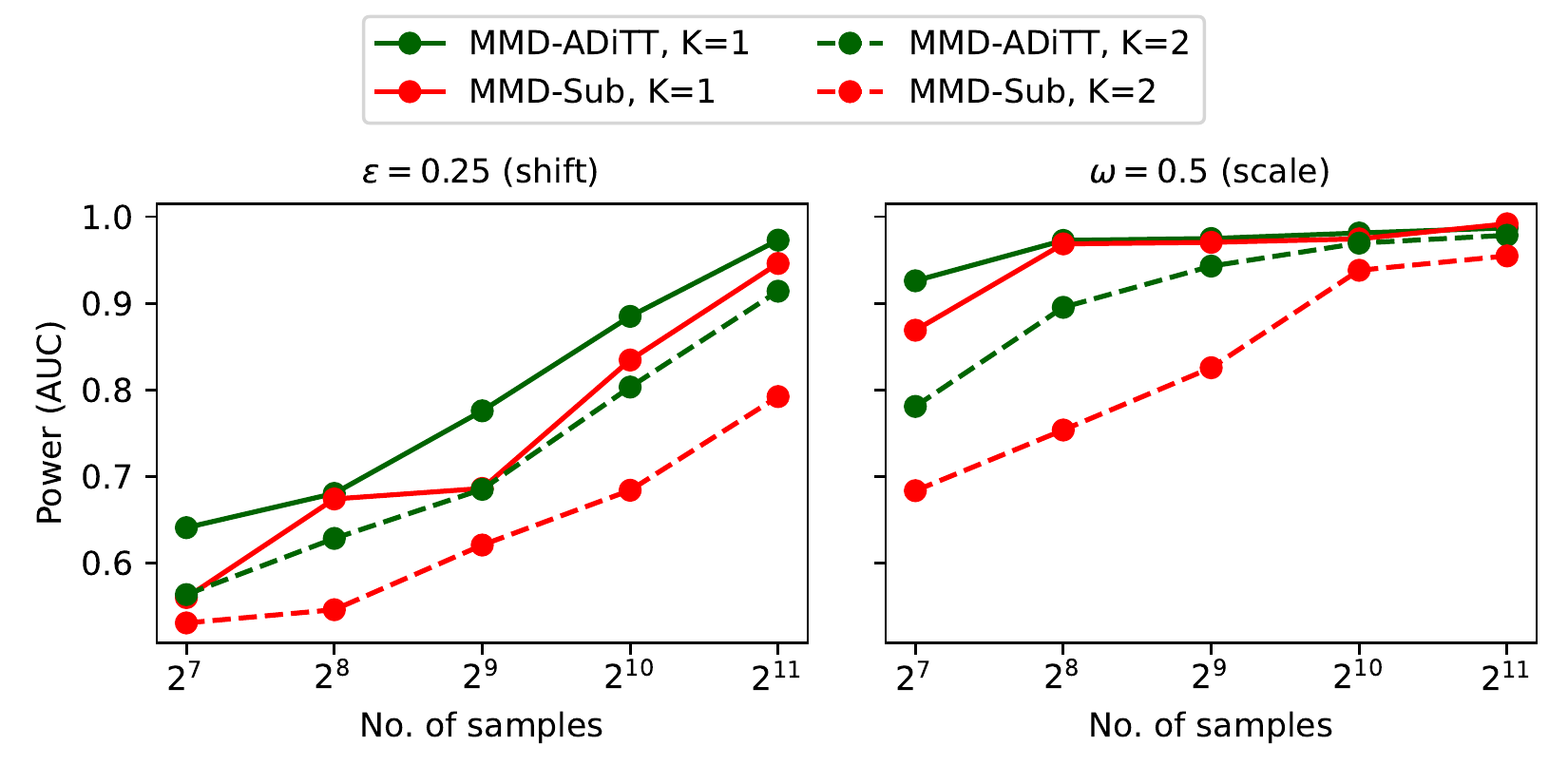}
    \vspace{-1mm}
    \caption{Plots showing how powerfully detectors respond when the context-conditional distribution $P_{S_1|C}$ changes from $N(C,1)$ to $N(C+\epsilon,\omega^2)$. We consider both unimodal ($K=1$) and bimodal ($K=2$) deployment context distributions $P_{C_1}$ and in the latter case only change $P_{S_1|C}$ for one mode.}
    \label{fig:tv_power}
    \end{center}
    \vspace{-1mm}
\end{figure}

% We make the problem more challenging with deployment context distribution corresponding to a mixture of Gaussians $P_{C_1}=\sum_{k=1}^K N(\mu_k,\sigma_k^2)$. For each of the 100 runs we generate new means $\{\mu_k\}_{k=1}^K$ (from a $N(0,1)$) and fix $\sigma=0.2$, resulting in deployment samples such as that shown in Figure~\ref{fig:tv_marginal_weights}. Figure~\ref{fig:tv_calib} confirms that the calibration of both the MMD-ADiTT and MMD-Sub methods is unaffected by the complexity of the change in the underlying context distribution.
To compare how powerfully detectors respond to changes in the context-conditional distribution we change $P_{S_1|C}$ from $S_1 \sim N(C,1)$ to $S_1 \sim N(C+\epsilon,\omega^2)$ for instances within one of the $K$ deployment modes, with an example for $K=2$ and $(\epsilon,\omega)=(0,2)$ shown in Figure~\ref{fig:tv_marginal_weights}. Figure~\ref{fig:tv_power} demonstrates how the power of each detector varies with sample size for the $K=1,2$ and $(\epsilon,\omega)=(0.25,0),(0,0.5)$ cases. 
We see that even for the unimodal case, where we might not necessarily expect the MMD-ADiTT approach to have an advantage, it is more powerful across all sample sizes and distortions considered (see Appendix~\ref{app:4_1} for more results). For the bimodal case the difference in performance is much larger. This is for an important reason that we illustrate by considering the difference between reference and deployment distributions shown in Figure~\ref{fig:tv_marginal_weights}.
Although MMD-Sub subsamples reference instances corresponding to the deployment modes to achieve a marginal weighting effect similar to that shown in Figure~\ref{fig:tv_marginal_weights}, the MMD is computed in a manner that weights the similarities between instances in different modes equally to the similarities between instances in the same mode. This adds noise to the test statistic that makes it more difficult to observe differences of interest.
% We could consider this as MMD computed using the full reference set where $W_{1,1}$ is uniformly equal to $(n_1)^{-2}$, $W_{0,1}$ has rows uniformly equal to either $(n'_0n_1)^{-1}$ or 0 and $W_{0,0}$ has rows and columns uniformly equal to either... 
Instead the ADiTT method leverages the information provided by the context, only comparing the similarity of instances with similar contexts. This can be observed by visualising the weights matrices $\textbf{W}_{0,0}$, $\textbf{W}_{1,1}$ and $\textbf{W}_{0,1}$ where the rows and columns are ordered by increasing context, as shown in Figure~\ref{fig:tv_heatmaps}. Appendix~\ref{app:4_1} gives further explanation and visualises the corresponding matrices for MMD-Sub. In summary, MMD-ADiTT detects differences conditional on context, i.e. differences between $P_{S_1|C}$ and $P_{S_0|C}$, whereas subsampling detects differences between the marginal distributions $P_{S_1}$ and $\int \text{d}c P_{S_0|C}(\cdot|c)P_{C_1}(c)/P_{C_0}(c)$. 

\subsection{Controlling for subpopulation prevalences}

The population underlying reference data can often be decomposed into subpopulations between which the distributions underlying features, labels or their relationship may differ. Figure~\ref{fig:day-night} is one example where the distribution underlying both the image and its label differed depending on whether it corresponded to day or night. Often practitioners wish to be be alerted to changes in distributions underlying subpopulations, but not to changes in their prevalence. Sometimes subpopulation membership will not be available explicitly but will have to be inferred. If subpopulations are known and labels can be assigned to all (resp. some) of the reference data, a classifier could be trained in a fully (resp. semi-) supervised manner to map instances onto a vector representing subpopulation membership probabilities. Alternatively a fully unsupervised approach could be taken where a probabilistic clustering algorithm is used to identify subpopulations and map onto probabilities accordingly, which we demonstrate in the following experiments. 

We take $\mathcal{S}=\mathbb{R}^2$ and the reference distribution to be a mixture of two Gaussians. We wish to allow changes to the mixture weights but detect when a component is scaled by a factor of $\omega$ or shifted by $\epsilon$ standard deviations in a random direction. We do not assume access to subpopulation (component) labels and therefore train a Gaussian mixture model to associate instances with a probability that they belong to subpopulation 1, which is then used as context $C$. 

\begin{figure}
    \begin{subfigure}{0.49\linewidth}
      \includegraphics[trim={5mm 5mm 3mm 0mm}, clip, width=1.0\linewidth]{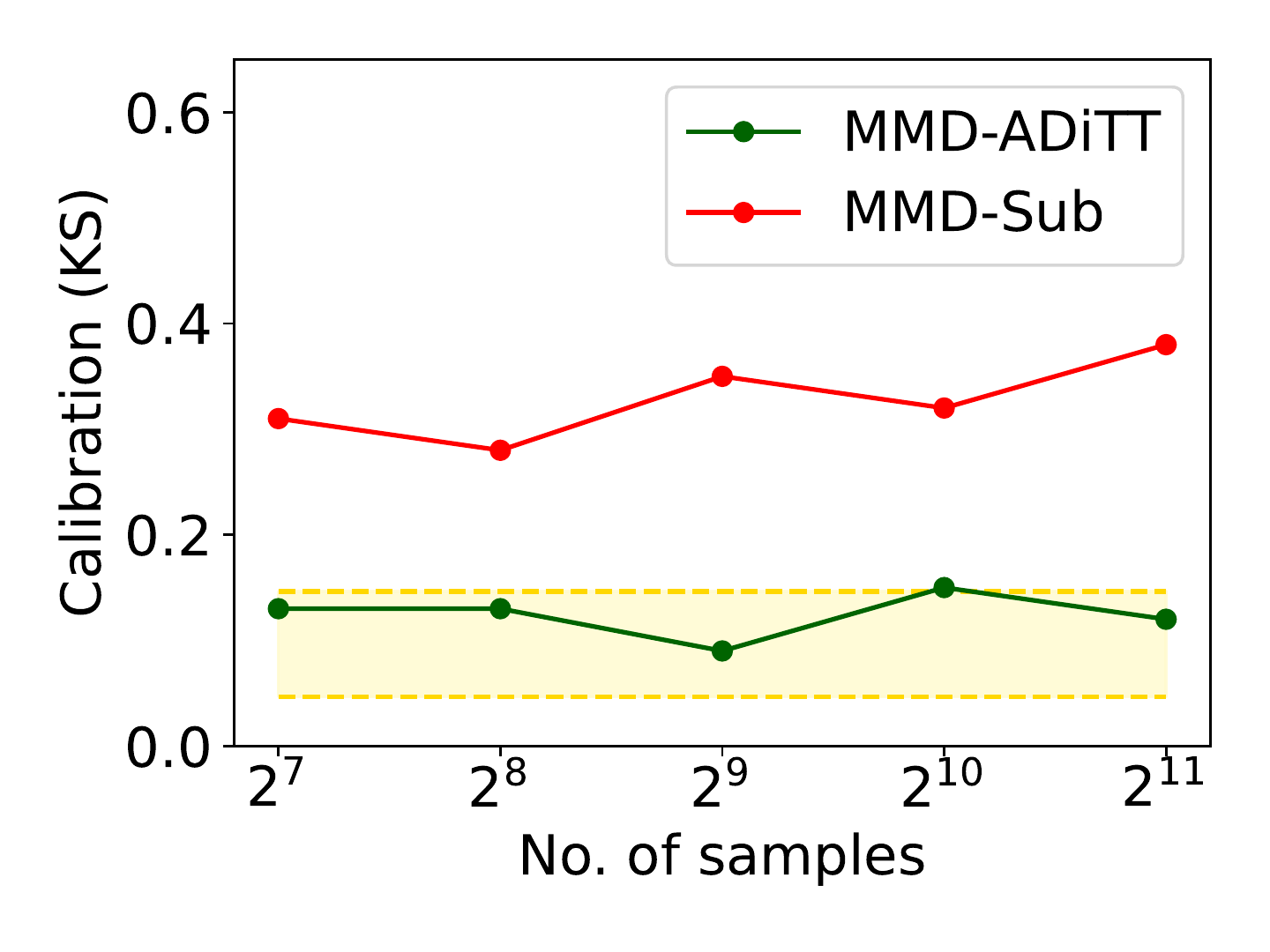}
    %   \caption{}
      \label{fig:blobs_calib}
    \end{subfigure}
    \begin{subfigure}{0.49\linewidth}
      \centering
      \includegraphics[trim={5mm 5mm 3mm 0mm}, clip, width=1.0\linewidth]{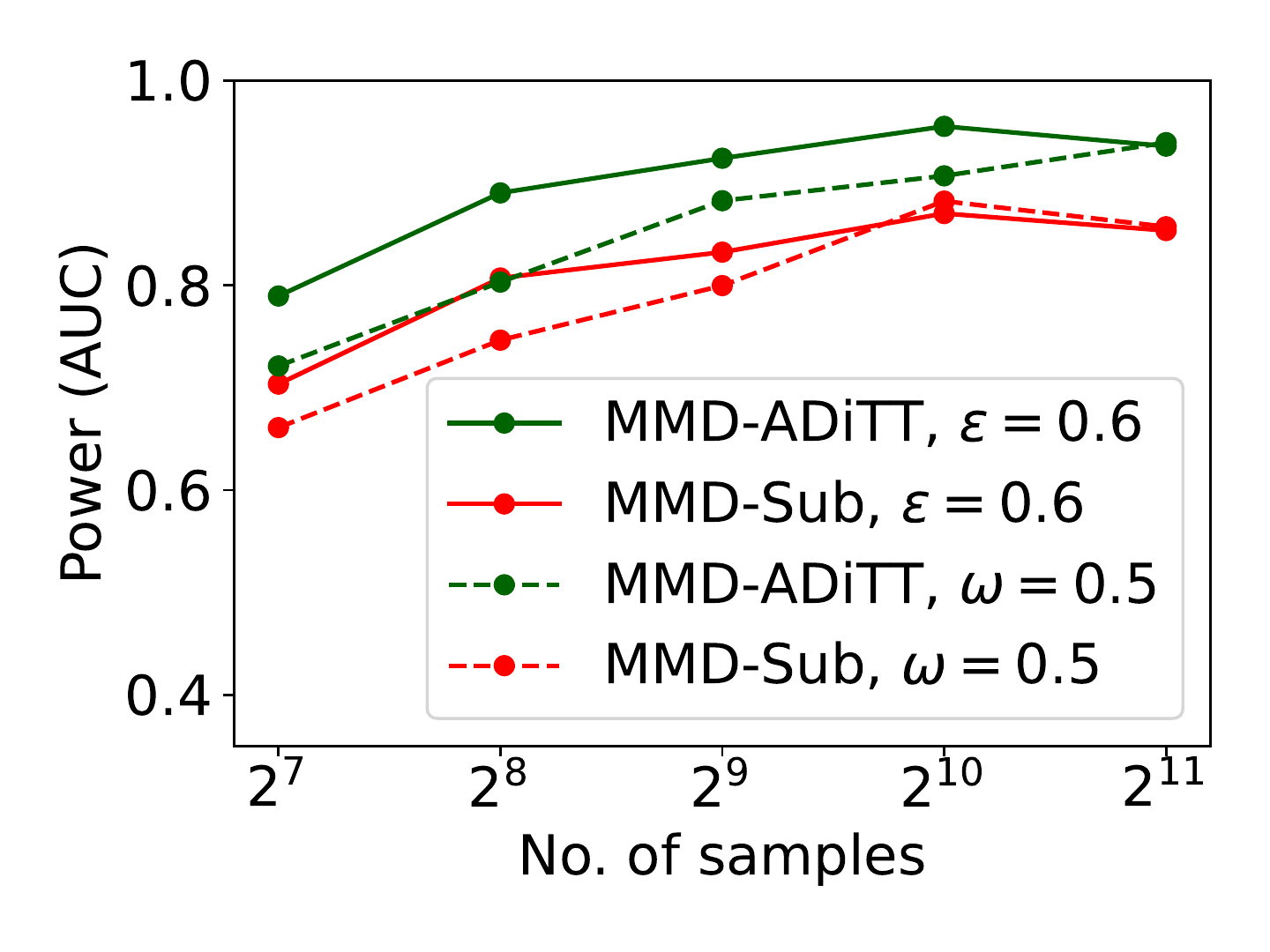}
    %   \caption{}
      \label{fig:blobs_data_nulls}
    \end{subfigure}
    \vspace{-2.5mm}
        \caption{Plots showing (left) calibration under resampling of subpopulation prevalences and (right) power under changes to a subpopulation mean ($\epsilon=0.6$) or scale ($\omega=0.5$).}
      \label{fig:blobs_power_calib}
       \vspace{-1mm}
    \end{figure}

Figure \ref{fig:blobs_power_calib} shows how the calibration, and -- for the $\epsilon=0.6$ and $\omega=0.5$ cases -- power, varies with sample size. Given that the MMD-ADiTT method remains well calibrated, we can assume the miscalibration of the MMD-Sub detector results from the bias resulting from imperfect density estimation used for rejection sampling. Although the fit will improve with the number of samples, it seems to be outpaced by the increase in power with which the bias is detected as the sample size grows. Moreover we see that across all sample sizes MMD-ADiTT more powerfully detects both changes in mean and variance, with additional results shown in Appendix~\ref{app:4_1}.

% Consider the reference distribution $S_0 \sim \pi_1N(\mu_1,\sigma_1^2I)+\pi_2N(\mu_2,\sigma^2_2I)$, where $\mu_1,\mu_2\in\mathbb{R}^2$ and $\sigma_1,\sigma_2\in\mathbb{R}$, containing two subpopulations with proportions $\pi_1$ and $\pi_2$. We do not assume access to subpopulation labels and therefore train a Gaussian mixture model to associate instances with a probability that they belong to subpopulation 1. This probability is then used as the context variable $C$ to detect drift in a manner that is sensitive to changes in $(\mu_1,\sigma_1,\mu_2,\sigma_2)$ but insensitive to changes in $(\pi_1, \pi_2)$. For each test run we generate reference proportions from a $\text{Beta}(2,2)$ distribution and deployment proportions from a $\text{Beta}(1,1)$, such that deployment samples are more likely to be heavily dominated by a single subpopulation. As alternatives we either shift the Gaussian distribution underlying the dominant class by $\epsilon\in\{0.2,0.4,0.6,0.8,1.0\}$ standard deviations in a random direction or scale its standard deviation by a factor of $\lambda\in\{0.5,2\}$. Table~\ref{tab:blobls_results} lists, for the $n_0=n_1=1024$ case, the calibration under the null as well as the power under the various alternatives. Figure \ref{fig:blobs_power_calib} shows how the calibration, and -- for the $\epsilon=0.6$ and $\lambda=0.5$ cases -- power, varies with sample size. Results for all combinations of distortions and sample size can be found in the Appendix.

    % \begin{table*}
    %     \centering
    %     \caption{\vspace{3mm}
    %     }\label{tab:blobls_results}
    %     \small
    %         \begin{tabular}{| *{9}{c|} }
    %             \hline
    %             Method & Calibration (KS) & \multicolumn{7}{c|}{Power (AUC)} \\ 
    %          \hline
    %          &  & $\epsilon=0.2$ & $\epsilon=0.4$ & $\epsilon=0.6$ & $\epsilon=0.8$ & $\epsilon=1.0$ & $\lambda=0.5$ & $\lambda=2.0$\\
    %          \hline
    %          $\text{MMD-ADiTT}$ & \textbf{0.14} & \textbf{0.76} & \textbf{0.93} & \textbf{0.96} & \textbf{0.97} & \textbf{0.94}& \textbf{0.97} & \textbf{0.97}\\
    %          $\text{MMD-Sub}$ & 0.35 & 0.65 & 0.81 & 0.87 & 0.87 & 0.88 & 0.86 & 0.87\\
    %         \hline
    %     \end{tabular}
    % \end{table*}

\subsection{Controlling for model predictions}

% Recall that the unavailability of labels during deployment usually limits practitioners to the detection of covariate drift. Many choose to look for covariate drift in the space of model predictions, the idea being that drift into unfamiliar regions of covariate space will manifest as more frequent underconfident or underconfident predictions. 
Detecting covariate drift conditional on model predictions allows a model to be more or less confident than average on a batch of deployment samples, or more frequently predict certain labels, as long as the features are indistinguishable from reference features for which the model made similar predictions. Covariate drift into a mode existing in the reference set would therefore be permitted whereas covariate drift into a newly emerging concept would be detected.
% Note here that this approach also corresponds to aiming to allow subpopulation shift where the subpopulations correspond to the class labels being predicted. In this case testing for change in the conditional distribution $P_{Y|X}$ corresponds to testing whether the "label shift assumption" holds, which is a crucial assumption made by many approaches to domain adaptation.
We use the ImageNet \citep{deng2009imagenet} class structure developed by \citet{santurkar2021breeds} to represent realistic drifts in distributions underlying subpopulations. The ImageNet classes are partitioned into 6 semantically similar superclasses and drift in the distribution underlying a superclass corresponds to a change to the constituent subclasses.
Adhering to the popularity of self-supervised backbones in computer vision we define a model $M(x)=H(B(x))$, where $B$ is the convolutional base of a pretrained SimCLR model \citep{chen2020simple} and $H$ is a classification head we train on the ImageNet training split to predict superclasses. Experiments are then performed using the validation split. 
% We additionally use the pretrained SimCLR model as part of the kernel $k:\mathcal{X}\times\mathcal{X}\rightarrow\mathbb{R}$, applying the base $B$ followed by the projection head $P$ to images before applying the usual Gaussian RBF kernel in $\mathbb{R}^{128}$.
We also use the pretrained SIMCLR model as part of the kernel $k$, applying both the base and projection head to images before applying the usual Gaussian RBF in $\mathbb{R}^{128}$.

\begin{figure}
    \begin{subfigure}{0.49\linewidth}
      \includegraphics[trim={5mm 5mm 3mm 0mm}, clip, width=1.0\linewidth]{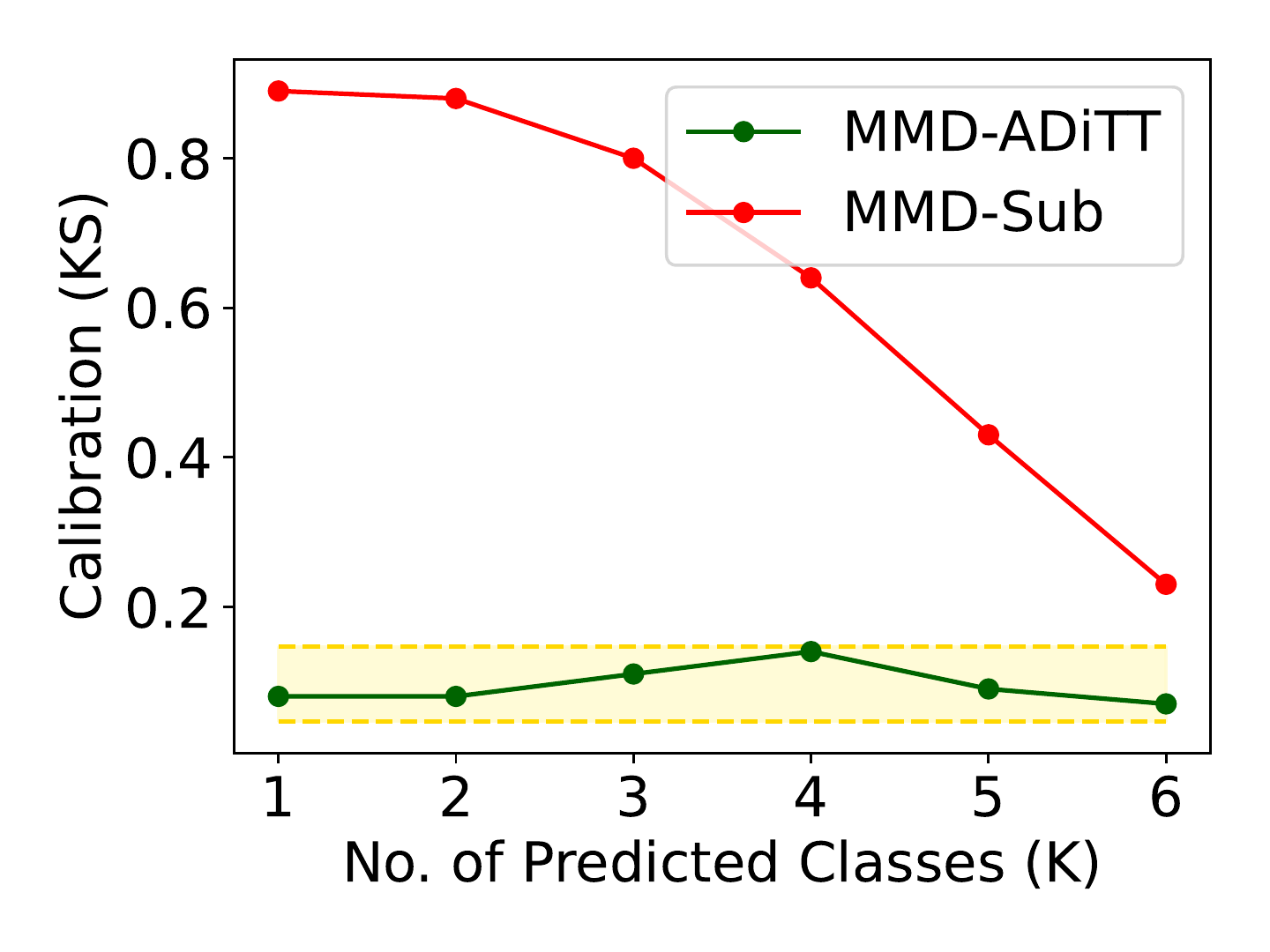}
    %   \caption{\small{.}}
      \label{fig:breeds_calib}
    \end{subfigure}
    \begin{subfigure}{0.49\linewidth}
      \centering
      \includegraphics[trim={5mm 5mm 3mm 0mm}, clip, width=1.0\linewidth]{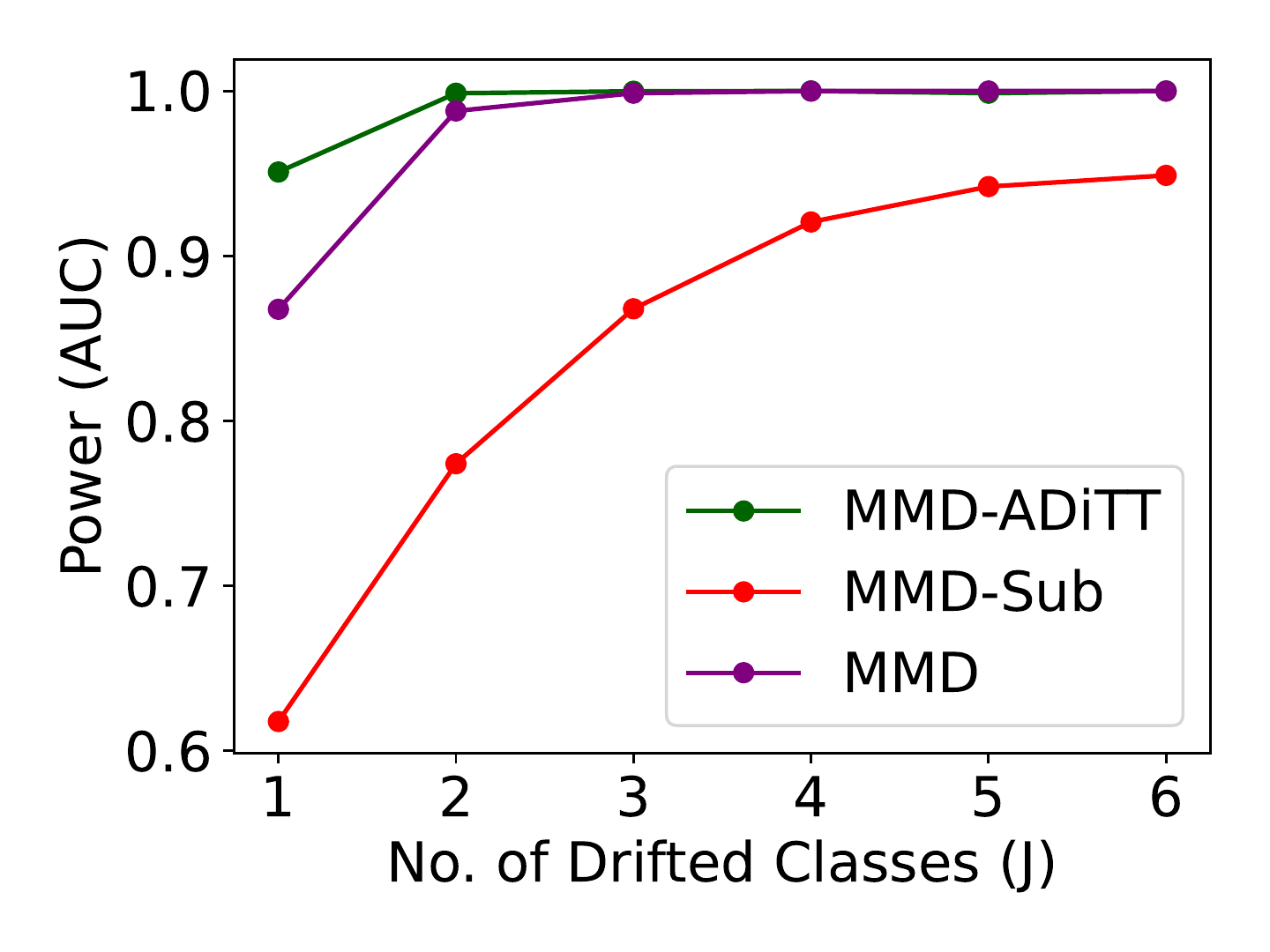}
    %   \caption{\small{.}}
      \label{fig:breeds_power}
    \end{subfigure}
     \vspace{-2.5mm}
        \caption{Plots showing (left) calibration under changes to model predictions resulting in only $K$ of 6 predicted classes and (right) power under changes to $J$ of 6 class distributions.}
      \label{fig:breeds_power_calib}
      \vspace{-1mm}
    \end{figure}
    
To investigate calibration under changing context $C=M(X)$, we randomly choose $K\in\{1,..,6\}$ of the 6 labels and sample a deployment batch containing only images to which the model assigns a most-likely label within the $K$ chosen. Therefore for $K=6$ the marginal distribution of the images remains the same in both the reference and deployment sets, but for $K<6$ they differ in a way we wish to allow. In Figure~\ref{fig:breeds_power_calib} we see that MMD-Sub was unable to remain calibrated for this trickier problem requiring density estimators to be fit in six-dimensional space. By contrast the MMD-ADiTT method remains well calibrated. Given MMD-Sub's ineffectiveness on this harder problem, which can be further seen in Figure~\ref{fig:breeds_power_calib}, we provide an alternative baseline to contextualise power results. The standard MMD two-sample test does not allow for changes in distribution that can be attributed to changes in model predictions and therefore does not have built in insensitivities like MMD-ADiTT. We might therefore expect it to respond more powerfully to drift generally, including those corresponding to specific subpopulations. Figure~\ref{fig:breeds_power_calib} shows however that this is not generally the case. When only one or two subpopulations have drifted the MMD-ADiTT detector, by focusing on the type of drift of interest, is able to respond more powerfully. As the number of drifted subpopulations increases to $J\geq3$, such that the distribution has changed in a more global manner, the standard MMD test is equally powerful, as might be expected. Table~\ref{tab:blobs_results} in Appendix~\ref{app:4_3} shows that this pattern holds more generally.

% We were unable to implement MMD-Sub as a baseline for these experiments due to difficulty fitting a kernel density estimator in $\mathbb{R}^6$ where many instances get assigned identical pre

% Figure~\ref{fig:breeds_calib} shows that when the deployment batch only contains samples for which 
\section{Conclusion}

We introduced a new framework for drift detection which breaks from the i.i.d.\ assumption by allowing practitioners to specify context under which the deployment data is permitted to change.
This drastically expands the space of problems to which drift detectors can be usefully applied. In future work we intend to further explore how certain combinations of contexts and statistics may be used to target certain types of drift, such as covariate drift into regions of high epistemic uncertainty.

% We introduced a new drift detection framework which breaks with the i.i.d.\ assumption by allowing practitioners to specify context under which the deployment data is permitted to change. This makes drift detection practical for many real-world applications. We are especially interested to further explore different types of context variables as well as integrating various (learned) test statistics in our framework. 

% 1. summarize -> overcome iid assumption, make drift detection practical in real world
% 2. shown desired behaviour (powerful yet well calibrated) for various problems
% 3. interested to see how it will be used for: (a) different types of conditioning, (b) different test statistics, e.g. learned, and (c) different data modalities

% We introduce a new framework for drift detection which breaks from the i.i.d.\ assumption rendering existing approaches inapplicable to many problems. We instead afford practitioners the flexability to specify context 

\section*{Acknowledgements}

We would like to thank Ashley Scillitoe and Hao Song for their help integrating our research into the open-source Python library \texttt{alibi-detect}.

\bibliography{main}
\bibliographystyle{icml2022}

% Comment out whilst iterating
%%%%%%%%%%%%%%%%%%%%%%%%%%%%%%%%%%%%%%%%%%%%%%%%%%%%%%%%%%%%%%%%%%%%%%%%%%%%%%%
%%%%%%%%%%%%%%%%%%%%%%%%%%%%%%%%%%%%%%%%%%%%%%%%%%%%%%%%%%%%%%%%%%%%%%%%%%%%%%%
% APPENDIX
%%%%%%%%%%%%%%%%%%%%%%%%%%%%%%%%%%%%%%%%%%%%%%%%%%%%%%%%%%%%%%%%%%%%%%%%%%%%%%%
%%%%%%%%%%%%%%%%%%%%%%%%%%%%%%%%%%%%%%%%%%%%%%%%%%%%%%%%%%%%%%%%%%%%%%%%%%%%%%%
\newpage
\appendix\label{appendix}
\onecolumn

\section{Setting Regularisation Parameters for the MMD-based ADiTT Estimator}\label{add:lam_tune}

Consider the problem of estimating the CoDiTE function $U_\text{MMD}:\mathcal{C}\rightarrow \mathbb{R}$, defined in Equation~\ref{eqn:mmd-codite} as
\begin{equation}
    U_\text{MMD}(c)=||\mu_{S_0|C=c}-\mu_{S_1|C=c}||^2_{\mathcal{H}_k},
\end{equation}
where $\mu_{S_z|C=c}$ is the kernel mean embedding of $P_{S^z|C=c}$, which by unconfoundedness is equal to $P_{S_z|C}:=P_{S|C,Z=z}$. Unconfoundedness allows us to use the methods of \citet{park2020measure} on samples $\{(s^0_i,c^0_i)\}_{i=1}^{n_0}$ and $\{(s^1_i,c^1_i)\}_{i=1}^{n_1}$ separately to obtain estimators $\hat{\mu}_{S_0|C=c}$ and $\hat{\mu}_{S_1|C=c}$ of CMEs $\mu_{S_0|C}$ and $\mu_{S_1|C}$ respectively. \citet{park2021conditional} show that the plug-in estimator
\begin{equation}
    \hat{U}_\text{MMD}(c)=||\hat{\mu}_{S_0|C=c}-\hat{\mu}_{S_1|C=c}||^2_{\mathcal{H}_k},
\end{equation}
for which a closed form expression is given in Equation~\ref{eqn:mmd-codte-est}, is then consistent in the sense that 
\begin{equation}
    \mathbb{E}[(\hat{U}_\text{MMD}(C)-U_\text{MMD}(C))^2]\xrightarrow{P} 0 \;\;\;\text{as}\;\;\; n_0,n_1\rightarrow \infty.
\end{equation}

\citet{park2020measure}'s method for estimating the CME $\mu_{S_0|C}$ from samples $\{(s^0_i,c^0_i)\}_{i=1}^{n_0}$ first considers the RKHS $\mathcal{G}_{\mathcal{S}\mathcal{C}}$ of functions from $\mathcal{C} \rightarrow \mathcal{H}_k$ induced by the operator-valued kernel $l_{\mathcal{S}\mathcal{C}}=l(z,z')\text{Id}$ where $l:\mathcal{C} \times \mathcal{C} \rightarrow \mathbb{R}$ is a kernel on $\mathcal{C}$ and $\text{Id}:\mathcal{H}_k \rightarrow \mathcal{H}_k$ is the identity operator. If one then uses $\mathcal{G}_{\mathcal{S}\mathcal{C}}$ as the hypothesis space for a regression of the functions $\{k(s^0_i,\cdot)\}_{i=1}^{n_0}$ against contexts $\{c^0_i\}_{i=1}^{n_0}$ under the regularised objective 
\begin{equation}
    \frac{1}{n_0}\sum_{i=1}^{n_0}||k(s_i^0,\cdot)-f(c_i^0)||_{\mathcal{H}_k}^2 + \lambda_0 ||f||_{\mathcal{G}_{\mathcal{S}\mathcal{C}}}^2,
\end{equation}
then a representer theorem applies that states there exists an optimal solution of the form
\begin{equation}
    f_{\lambda_0}(c)=\sum_i\alpha_i l(c,c^0_i)=\boldsymbol{\alpha}^{\top}\mathbf{l}_0(c),
\end{equation}
where
\begin{equation}
    \boldsymbol{\alpha}=(\mathbf{L}_{0,0}+n_0\lambda_0 \mathbf{I})^{-1}\mathbf{k}_0(\cdot)\in\mathcal{H}_k^{n_0}.
\end{equation}
We therefore recommend running an optimisation process for $\lambda_0$ which for each candidate value splits $\{(s^0_i,c^0_i)\}_{i=1}^{n_0}$ into $k$-folds, computes $f_{\lambda_0}(c)$ for each fold, and sums the squared errors $||k(s,\cdot)-f_{\lambda_0}(c)||_{\mathcal{H}_k}^2$ across out-of-fold instances $(s,c)$. Recalling the shorthand $\mathbf{L}_{\lambda_0}^{-1}=(\mathbf{L}_{0,0}+n_0\lambda_0 \mathbf{I})^{-1}$, this can be achieved by noting 
\begin{align}
    ||k(s,\cdot)-f_{\lambda_0}(c)||_{\mathcal{H}_k}^2
    =&||k(s,\cdot) - \mathbf{l}_0(c)\mathbf{L}_{\lambda_0}^{-1}\mathbf{k}_0(\cdot)||_{\mathcal{H}_k}^2 \\
    =& \langle k(s,\cdot),k(s,\cdot)\rangle_{\mathcal{H}_k} + \langle\mathbf{l}_0(c)\mathbf{L}_{\lambda_0}^{-1}\mathbf{k}_0(\cdot),\mathbf{l}_0(c)\mathbf{L}_{\lambda_0}^{-1}\mathbf{k}_0(\cdot)\rangle_{\mathcal{H}_k} -2\langle \mathbf{l}_0(c)\mathbf{L}_{\lambda_0}^{-1}\mathbf{k}_0(\cdot),k(s,\cdot)\rangle_{\mathcal{H}_k} \\
    =& \langle k(s,\cdot),k(s,\cdot)\rangle_{\mathcal{H}_k} + \langle \sum_{i,j}l(c_i,c)(\mathbf{L}_{\lambda_0}^{-1})_{i,j}k(s_j,\cdot),\sum_{u,v}l(c_u,c)(\mathbf{L}_{\lambda_0}^{-1})_{u,v}k(s_v,\cdot)\rangle_{\mathcal{H}_k} \\ &-2\langle\sum_{i,j}l(c_i,c)(\mathbf{L}_{\lambda_0}^{-1})_{i,j}k(s_j,\cdot),k(s,\cdot)\rangle_{\mathcal{H}_k} \\ 
    =& k(s,s)+ \sum_{i,j,u,v}l(c_i,c)(\mathbf{L}_{\lambda_0}^{-1})_{i,j}k(s_j,s_v)l(c_u,c)(\mathbf{L}_{\lambda_0}^{-1})_{u,v} -2\sum_{i,j}l(c_i,c)(\mathbf{L}_{\lambda_0}^{-1})_{i,j}k(s_j,s) \\ 
    =& 
    k(s,s) + \mathbf{l}_0(c)\mathbf{L}_{\lambda_0}^{-1}\textbf{K}_{0,0}\mathbf{L}_{\lambda_0}^{-\top}\mathbf{l}_0(c)^{\top} - 2\mathbf{l}_0(c)\mathbf{L}_{\lambda_0}^{-1}\textbf{k}_0(s)^{\top}.
    \end{align}
The errors for all out-of-fold instances can be computed in one go by stacking $\mathbf{l}_0(c)$ and $\textbf{k}_0(s)$ vectors into matrices. The same procedure can then be performed to select $\lambda_1$.

\section{Implementation Details for Drift Detection with the MMD-based ADiTT Estimator}\label{app:aditt}

In this section we make clear the exact process used for computing the MMD-ADiTT test statistic and estimating the associated p-value representing its extremity under the null hypothesis. Recall from Equations~\ref{eqn:aditt-est} and \ref{eqn:mmd-codte-est} that the test statistic is the average MMD-based CoDiTE estimate on a set of held-out deployment contexts $\tilde{\mathbf{c}}^1$. The portion of samples we hold out for this purpose is 25\% across all experiments. CoDiTE estimates require a regularisation parameter $\lambda$ to use as part of the estimation process, for which we use $\lambda=0.001$ across all experiments. For kernels $k: \mathcal{S} \times \mathcal{S} \rightarrow \mathbb{R}$ and $l: \mathcal{C} \times \mathcal{C} \rightarrow \mathbb{R}$ we use the Gaussian RBF $f(x,x')=\exp(-\frac{(x-x')^2}{2\sigma^2})$ where $\sigma$ is set to be the median distance between all reference and deployment statistics, for $k$, or contexts, for $l$.

To associate resulting test statistics with estimates of p-values we use the conditional permutation test of \citet{rosenbaum1984conditional}. We do so using $n_{\text{perm}}=100$ conditional permutations. This requires fitting a classifier $\hat{e}:\mathcal{C}\rightarrow[0,1]$ to approximate the propensity score $e(c)=P(Z=1|C=c)$. We do this by training a kernel logistic regressor on the data $\{(c_i,z_i)\}_{i=1}^n$, using the same kernel $l$ defined above. More precisely, we achieve this by first fitting a kernel support vector classifier (SVC) and then performing logistic regression on the scores to obtain probabilities. We found that mapping SVC scores onto probabilities in this manner using just two logistic regression parameters meant that overfitting to $\{(c_i,z_i)\}_{i=1}^n$ was not a problem.

\section{MMD-Sub Baseline}\label{app:baseline}

In this section we make clear the exact process used for computing the MMD-Sub test statistic and estimating the associated p-value representing its extremity under the null hypothesis. Recall from Section~\ref{sec:exps} that MMD-Sub aims to obtain a subset of reference instances for which the underlying context distribution matches $P_{C_1}$. It proceeds by first fitting kernel density estimators $\hat{P}_{C_0}$ and  $\hat{P}_{C_1}$ to approximate $P_{C_0}$ and $P_{C_1}$. We use 25\% of the reference and deployment contexts for fitting the estimators and then hold out the samples from the rest of the process. We again use Gaussian RBF kernels but this time allow the bandwidth to be tuned using Scott's Rule \cite{scott2015multivariate}. Once fit, we retain for each $i$ in the set of unheld indices $\mathcal{U}$, reference sample $i$ with probability $\hat{P}_{C_1}(c_i)/(m\hat{P}_{C_0}(c_i))$, where $m=\max_{j\in\mathcal{U}}\hat{P}_{C_0}(c_j)/(\hat{P}_{C_1}(c_j))$.

Once a subset of the reference set has been sampled, an MMD two-sample test is applied against the deployment set in the normal way \cite{gretton2012mmd}. We use the same Gaussian RBF kernel with median heuristic and estimate the p-value using a conventional (unconditional) permutation test, again with $n_{\text{perm}}=100$.

\section{MMD-ADiTE: An Ablation}\label{app:adite}

In Section~\ref{sec:framework} we noted the importance of defining a test statistic that considers the difference between conditional distributions only at contexts supported by the deployment context distribution $P_{C_1}$, rather than the more general $P_{C}$. When using MMD-based CoDiTE estimators this corresponded to averaging over held out deployment contexts, rather than both reference and deployment contexts. We refer to the estimator that would have resulted from averaging over both reference and deployment contexts as MMD-ADiTE. We show in Tables~\ref{tab:tv_power_1} and \ref{tab:tv_power_2} that on the experiments considered in Section~\ref{sec:4_1} using MMD-ADiTE as a test statistic results in a wildly miscalibrated detector, as would be expected. Further details on these experiments can be found in Appendix~\ref{app:4_1}.

\section{Experiments: Further Details and Visualisations}\label{sec:app_expperiments}

The general procedure we follow for obtaining results is as follows. For calibration we define reference and deployment distributions $P_{S_0|C}(s|c)P_{C_0}(c)$ and $P_{S_1|C}(s|c)P_{C_1}(c)$ respectively where $P_{S_0|C}(s|c)$ is equal to $P_{S_1|C}(s|c)$ but $P_{C_0}(c)$ is not necessarily equal to $P_{C_1}(c)$. A single run then involves generating batches of reference and deployment data and applying a detector to obtain a p-value, with permutation tests performed using 100 permutations. We perform 100 runs to obtain 100 p-values. We then report the Kolmogorov-Smirnov distance between the empirical CDF of the 100 p-values and the CDF of the uniform distribution on $[0,1].$ 

For experiments exploring power, $P_{S_0|C}$ additionally differs from $P_{S_1|C}$. In this case we perform 100 runs where this change is present and 100 runs where only the change in $P_{C}$ is present. The ROC curve then plots TPRs, computed using the first 100 p-values, against FPRs, computed using the second 100 p-values. The area under the ROC curve -- the AUC -- is then reported as a measure of power.

\subsection{Controlling for domain specific context}\label{app:4_1}

\begin{figure}
    \centering
    \begin{subfigure}{0.4\linewidth}
    \centering
      \includegraphics[trim={5mm 5mm 3mm 5mm}, clip, width=0.85\linewidth]{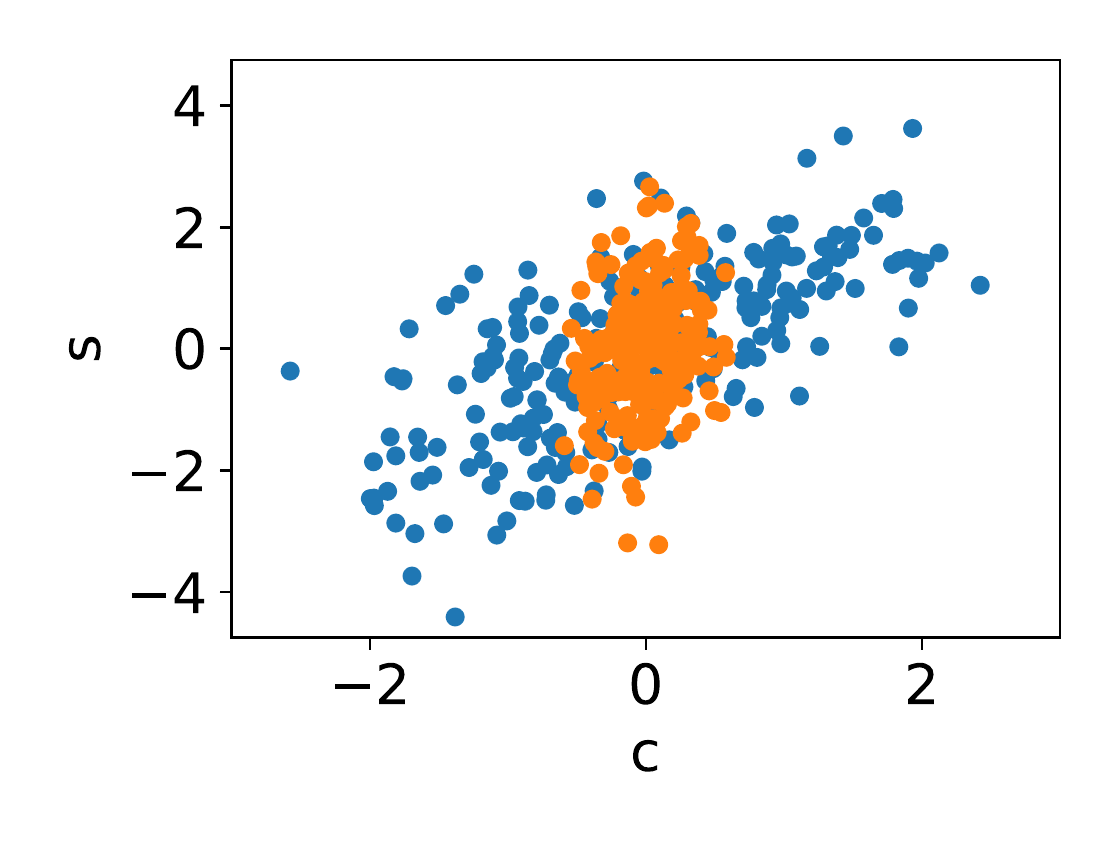}
      \caption{$C_1 \sim N(0,0.5^2)$.}
      \label{fig:tv_data_1}
    \end{subfigure}
    \begin{subfigure}{0.4\linewidth}
      \centering
      \includegraphics[trim={5mm 5mm 3mm 5mm}, clip, width=0.85\linewidth]{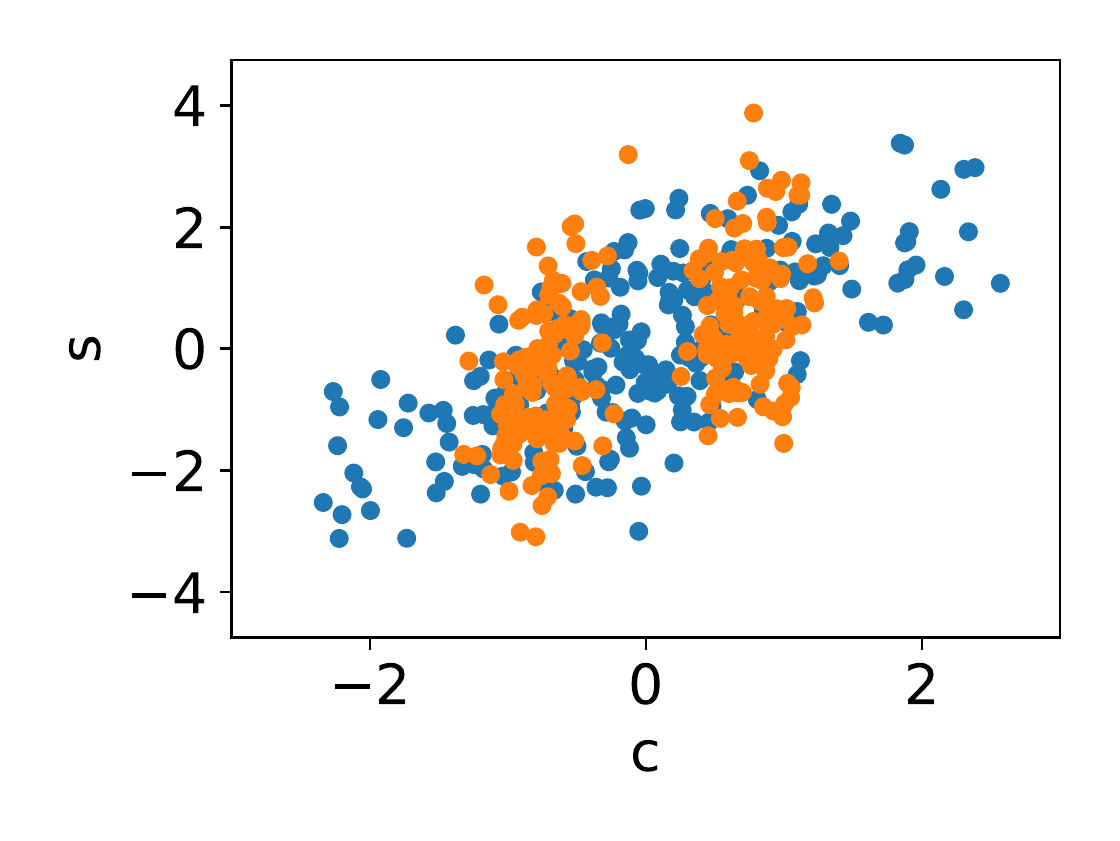}
      \caption{$C_1 \sim \frac12N(-0.75,0.2^2)+\frac12N(0.75,0.2^2)$}
      \label{fig:tv_data_2}
    \end{subfigure}
      \caption{Reference (blue) and deployment (orange) instances where the reference data has context-conditional distribution $S_0 \sim N(C_0,1)$ and context distribution $C_0 \sim N(0,1)$. The context-conditional distribution of the deployment instances remain the same, but the context distributions change to the distributions stated. We do not wish for these changes to result in a detection.}
      \label{fig:tv_data_12}
\end{figure}

For this problem we take $\mathcal{S}=\mathcal{C}=\mathbb{R}$. For the reference distribution we take $S_0| \sim N(C,1)$ and $C_0 \sim N(0,1)$, as shown in blue in both plots of Figures~\ref{fig:tv_data_12}, such that marginally $S_0 \sim N(0,2)$. We first consider a simple narrowing of the context distribution from $P_{C_0}=N(0,1)$ to $P_{C_1}=N(0,\sigma^2)$ for $\sigma\in\{0.125,0.25,0.5,1.0\}$ in order to demonstrate as clearly as possible how conventional two-sample tests fail to satisfy our notion of calibration, whereas MMD-ADiTT and MMD-Sub succeed.
The context-conditional distribution remains unchanged at $P_{S_0|C}=P_{S_1|C}=N(C,1)$. An example for the $\sigma=0.5$ case is shown in Figure~\ref{fig:tv_data_1} and the results are shown in Table~\ref{tab:tv_calib_narrow} for a sample size of $n_0=n_1=1000$.

\begin{figure}
    \centering
\includegraphics[width=0.8\linewidth]{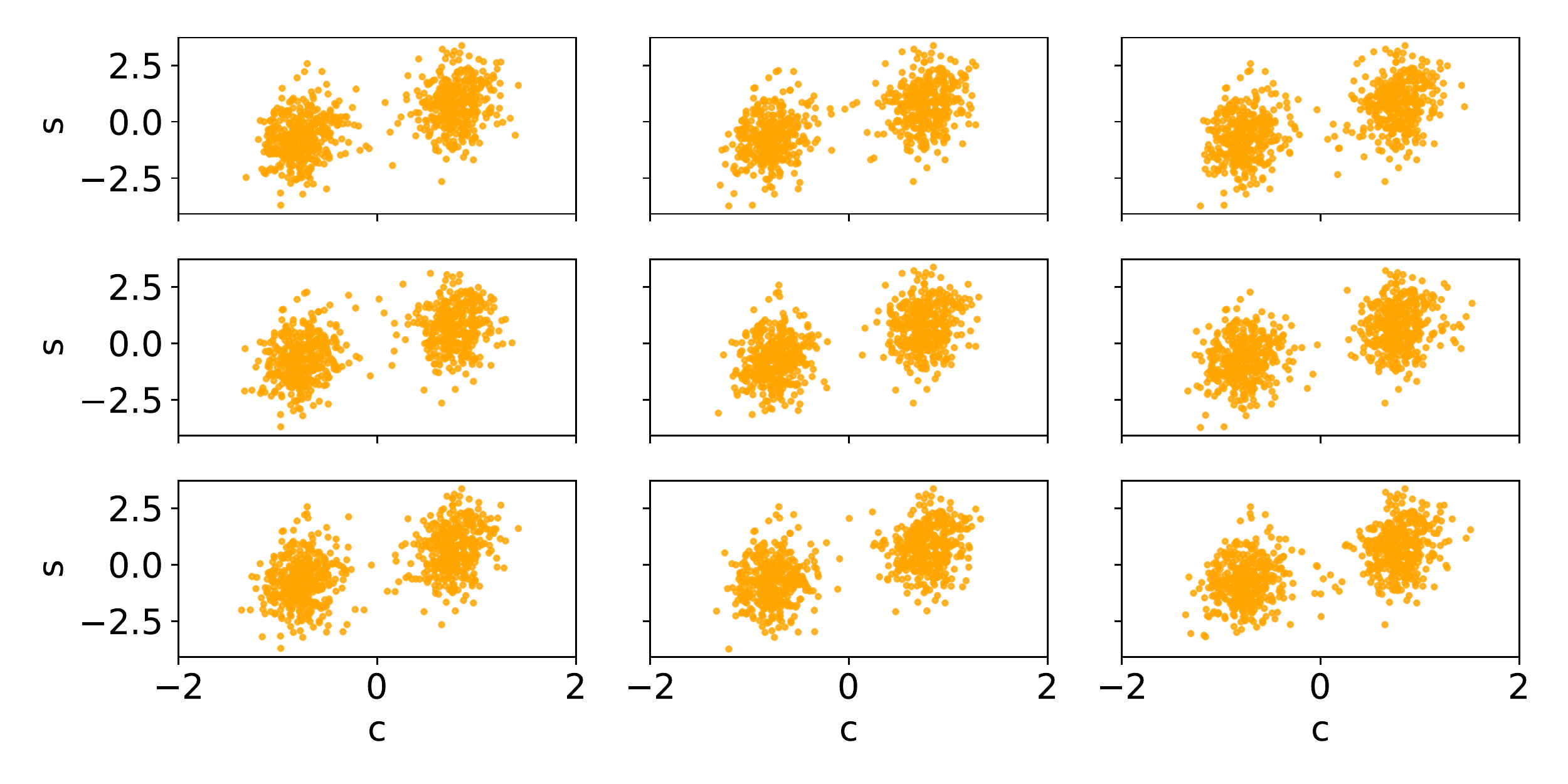}
\caption{The central plot shows here shows a batch of deployment samples shown generated under the same setup as Figure~\ref{fig:tv_data_2}. The surrounding plots all show alternative sets of reassigned deployment samples obtained by using the conditional resampling procedure of \citet{rosenbaum1984conditional} to reassign deployment statuses as $z'_i\sim\text{Ber}(e(c_i))$ for $i=1,...,n$. Note that the alternatives do not use identical samples for each reassignment, but do manage to achieve the desired context distribution, with none of the plots being noticeably different from any other.}
\label{fig:tv_perms}
\end{figure}

\begin{figure}
    \centering
\includegraphics[width=0.8\linewidth]{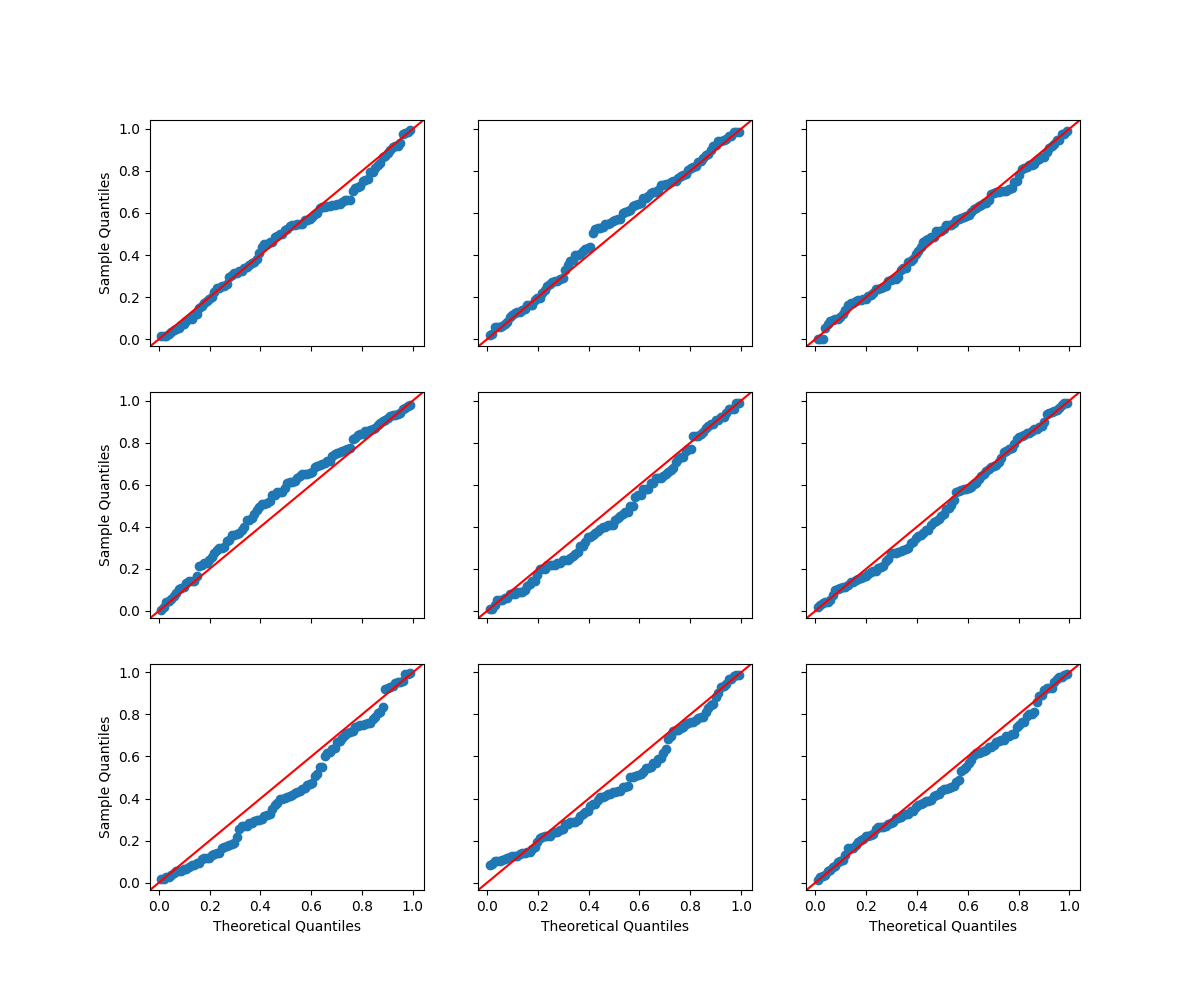}
\caption{Shown centrally is the Q-Q plot of a $U[0,1]$ against the p-values obtained by MMD-ADiTT under a change in context distribution from $N(0,1)$. The context-conditional distribution has not changed and therefore a perfectly calibrated detector should have a Q-Q plot lying close to the diagonal. To contextualise how well the central plot follows the diagonal, we surround it with Q-Q plots corresponding to 100 p-values actually sampled from $U[0,1]$.}
\label{fig:tv_qqs}
\end{figure}

Secondly to test whether detectors remain calibrated under more complex changes in context, we consider for the deployment context distribution a mixture of Gaussians $P_{C_1}=\frac{1}{K}\sum_{k=1}^K N(\mu_k,\sigma_k^2)$.
For each of the 100 runs we generate new means $\{\mu_k\}_{k=1}^K$ from a $N(0,1)$ and fix $\sigma_k=0.2$, resulting in deployment samples such as that shown in Figure~\ref{fig:tv_data_2} for the $K=2$ case. Again the context-conditional distribution remains unchanged at $P_{S_0|C}=P_{S_1|C}=N(C,1)$. Results are shown in Table~\ref{tab:tv_calib_complex} for a sample size of $n_0=n_1=1000$ and a Q-Q plot is shown in Figure~\ref{fig:tv_qqs} for the $K=2$ case. Figure ~\ref{fig:tv_perms} shows that the conditional resampling scheme indeed manages to reassign samples into reference and deployment windows in a diversity of ways whilst staying true the context distribution $P_{C_1}$.

\begin{figure}
    \centering
    \begin{subfigure}{0.4\linewidth}
    \centering
      \includegraphics[trim={5mm 5mm 3mm 5mm}, clip, width=0.85\linewidth]{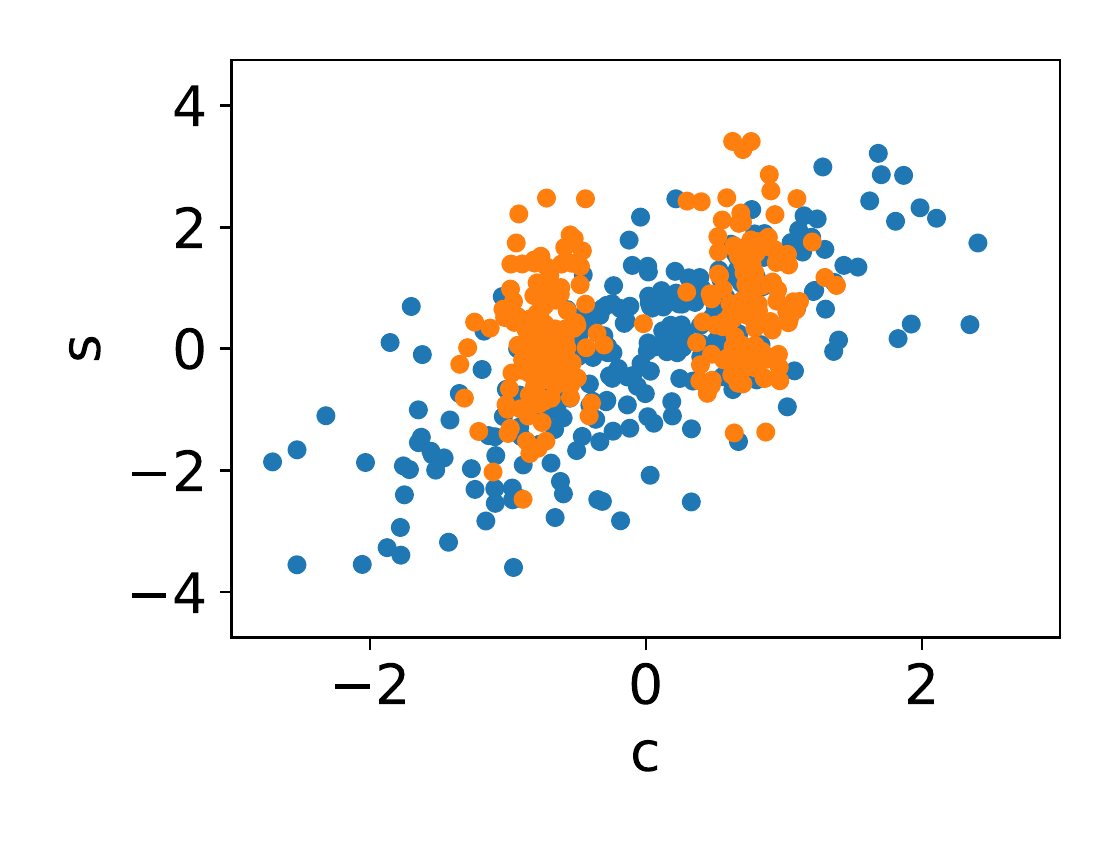}
      \caption{$(\epsilon, \omega)=(0.5,0)$}
      \label{fig:tv_data_3}
    \end{subfigure}
    \begin{subfigure}{0.4\linewidth}
      \centering
      \includegraphics[trim={5mm 5mm 3mm 5mm}, clip, width=0.85\linewidth]{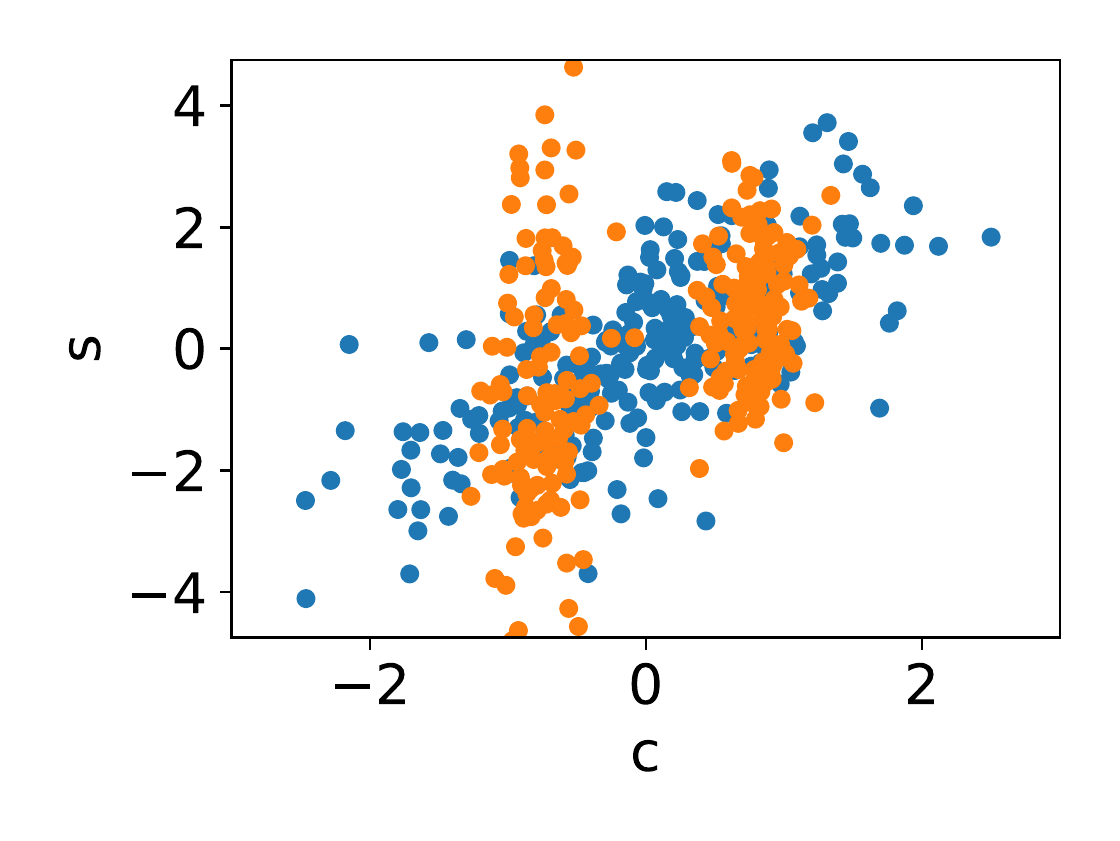}
      \caption{$(\epsilon, \omega)=(0,2)$}
      \label{fig:tv_data_4}
    \end{subfigure}
      \caption{Reference (blue) and deployment (orange) instances where the reference data has context-conditional distribution $S_0 \sim N(C_0,1)$ and context distribution $C_0 \sim N(0,1)$. The context distribution then changes $P_{C_1}=\frac12N(-0.75,0.2^2)+\frac12N(0.75,0.2^2)$ for the deployment sample and, for deployment instances corresponding to the first mode, the context-conditional distribution changes to $N(C+\epsilon,\omega^2)$  We wish for these changes to result in a detection.}
      \label{fig:tv_data_34}
\end{figure}

Finally to test power we again take the mixture of Gaussians deployment context distribution described above but now, for deployment instances in one of the $K$ modes, vary the context-conditional distribution from $N(C,1)$ to $N(C+\epsilon,\omega^2)$ for $\epsilon\in\{0.25,0.5\}$ or $\omega \in \{0.5,2.0\}$, with examples shown in Figure~\ref{fig:tv_data_34} for the $K=2$ case. Table~\ref{tab:tv_power_1} shows how the detectors' power varies with sample size for the unimodal ($K=1$) case. Table~\ref{tab:tv_power_2} shows the bimodal ($K=2$) case, where the difference in performance is more significant.

\begin{figure}
    \centering
    \includegraphics[width=0.75\linewidth]{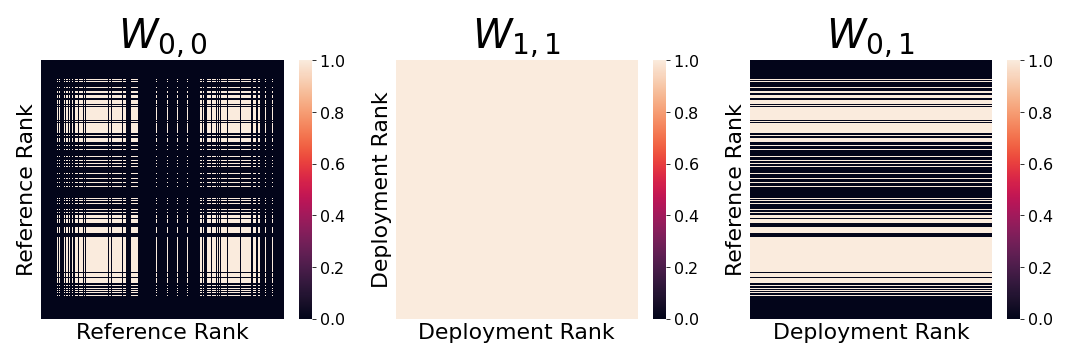}
    \caption{Visualisation of the weight matrices used in computation of the MMD-Sub when deployment contexts fall into two disjoint modes. We see that the similarities between instances in different modes of context contribute equally as the similarities between instances in the same mode of context.}
    \label{fig:tv_sub_heatmap}
\end{figure}

As noted in Section~\ref{sec:4_1}, this difference is because MMD-ADiTT detects differences conditional on context, i.e. differences between $P_{S_1|C}$ and $P_{S_0|C}$, whereas subsampling detects differences between the marginal distributions $P_{S_1}$ and $\int \text{d}c P_{S_0|C}(\cdot|c)P_{C_1}(c)/P_{C_0}(c)$. This was apparent from the block-structured weight matrices visualised in Figure~\ref{fig:tv_heatmaps}, corresponding to the context distributions of Figure~\ref{fig:tv_data_2}, showing that only similarities between instances with similar contexts significantly contributed. For example in $\mathbf{W}_{0,0}$ the blobs in the lower left corresponds to the weights assigned to similarities between reference instances whose contexts fall within the lower deployment context cluster and the blob in the upper right similarly corresponds to those within the upper cluster, with no weight assigned to similarities between the two. Figure~\ref{fig:tv_sub_heatmap} shows the corresponding matrices for MMD-Sub. Here rows and columns are fully active or inactive depending on whether a given reference instance was successfully rejection sampled. We again see similar blobs in the lower left and upper right, but now additionally in the upper left and lower right, adding unwanted noise to the test statistic. The pattern is even clearer for $\textbf{W}_{1,1}$ where, in the MMD-Sub case, all similarities between deployment instances are considered relevant, regardless of how similar their contexts are.

\begin{table*}
\small
    \centering
    \caption{Calibration under a narrowing of context from $N(0,1)$ to $N(0,\sigma^2)$}\label{tab:blobls_results}
    \label{tab:tv_calib_narrow}
    \vspace{2mm}
        \begin{tabular}{| *{10}{c|} }
            \hline
            Method  & \multicolumn{4}{c|}{Calibration (KS)} \\ 
         \hline
          & $\sigma=0.125$ & $\sigma=0.25$ & $\sigma=0.5$ & $\sigma=1.0$\\
         \hline
         $\text{MMD-ADiTT}$  & \textbf{0.10} & \textbf{0.06} & \textbf{0.09} & \textbf{0.09}\\
         $\text{MMD-Sub}$ & 0.12 & \textbf{0.06} & 0.16 & 0.19  \\
         $\text{MMD}$ & 1.00 & 1.00 & 0.96 & 0.10  \\
        \hline
    \end{tabular}
\end{table*}
\begin{table*}
\small
    \centering
    \caption{Calibration under a change of context from $N(0,1)$ to a mixture of Gaussians with $K$ components.}\label{tab:blobls_results}
    \label{tab:tv_calib_complex}
    \vspace{2mm}
        \begin{tabular}{| *{10}{c|} }
            \hline
            Method  & \multicolumn{5}{c|}{Calibration (KS)} \\ 
         \hline
          & $K=1$ & $K=2$ & $K=3$ & $K=4$ & $K=5$\\
         \hline
         $\text{MMD-ADiTT}$  & \textbf{0.06} & \textbf{0.11} & \textbf{0.11} & 0.14 & \textbf{0.08}\\
         $\text{MMD-Sub}$ & 0.11 & 0.19 & 0.12 & \textbf{0.11} & 0.14 \\
        \hline
    \end{tabular}
\end{table*}
\begin{table*}
\small
    \centering
    \caption{Power under a change of context-conditional distribution from $N(C,1)$ to $N(C+\epsilon,1)$ or $N(C,\omega^2)$.}\label{tab:tv_power_1}
    % \small
    \vspace{2mm}
        \begin{tabular}{| *{7}{c|} }
            \hline
            Method & Sample Size  & Calibration (KS) & \multicolumn{4}{c|}{Power (AUC)} \\ 
         \hline
         & & & $\epsilon=0.25$ & $\epsilon=0.5$ & $\omega=0.5$ & $\omega=2.0$\\
         \hline
         $\text{MMD-ADiTT}$ &  & 0.10 & \textbf{0.64} & \textbf{0.83} & \textbf{0.93} & \textbf{0.94}\\
         $\text{MMD-ADiTE}$ & 128 & 0.37 & 0.61 & 0.77 & 0.69 & 0.89\\
         $\text{MMD-Sub}$ &  & \textbf{0.08} & 0.56 & 0.77 & 0.87 & 0.87\\\hdashline
         $\text{MMD-ADiTT}$ &  & 0.14 & \textbf{0.68} & \textbf{0.89} & \textbf{0.97} & \textbf{0.98}\\
         $\text{MMD-ADiTE}$ & 256 & 0.52 & 0.63 & 0.75 & 0.77 & 0.92\\
         $\text{MMD-Sub}$ &  & \textbf{0.10} & 0.67 & 0.87 & \textbf{0.97} & 0.92\\\hdashline
         $\text{MMD-ADiTT}$ &  & \textbf{0.07} & \textbf{0.78} & \textbf{0.97} & \textbf{0.97} & \textbf{0.98}\\
         $\text{MMD-ADiTE}$ & 512 & 0.55 & 0.58 & 0.75 & 0.65 & 0.90\\
         $\text{MMD-Sub}$ &  & 0.10 &0.69 & 0.90 & \textbf{0.97} & 0.94\\\hdashline
         $\text{MMD-ADiTT}$ &  & 0.12 & \textbf{0.88} & \textbf{0.99} & \textbf{0.98} & \textbf{0.99}\\
         $\text{MMD-ADiTE}$ & 1024 & 0.50 & 0.62 & 0.71 & 0.55 & 0.88\\
         $\text{MMD-Sub}$ &  & \textbf{0.08} &0.83 & 0.98 & 0.97 & 0.98\\\hdashline
         $\text{MMD-ADiTT}$ &  & \textbf{0.09} & \textbf{0.97} & \textbf{0.99} & \textbf{0.99} & \textbf{0.99}\\
         $\text{MMD-ADiTE}$ & 2048 & 0.44 & 0.60 & 0.69 & 0.53 & 0.91\\
         $\text{MMD-Sub}$ &  & \textbf{0.09} & 0.95 & \textbf{0.99} & \textbf{0.99} & \textbf{0.99}\\
        \hline
    \end{tabular}
\end{table*}
\begin{table*}
\small
    \centering
    \caption{Power under a change of context-conditional distribution from $N(C,1)$ to $N(C+\epsilon,1)$ or $N(C,\omega^2)$ for instances in one of 2 deployment modes.}\label{tab:tv_power_2}
    \vspace{2mm}
        \begin{tabular}{| *{7}{c|} }
            \hline
            Method  & Sample Size  & Calibration (KS) & \multicolumn{4}{c|}{Power (AUC)} \\ 
         \hline
         & & & $\epsilon=0.25$ & $\epsilon=0.5$ & $\omega=0.5$ & $\omega=2.0$\\
         \hline
         $\text{MMD-ADiTT}$ &  & 0.09 & \textbf{0.56} & \textbf{0.73} & \textbf{0.78} & \textbf{0.87}\\
         $\text{MMD-ADiTE}$ & 128 & 0.22 & 0.55 & 0.69 & 0.66 & 0.84\\
         $\text{MMD-Sub}$ & & \textbf{0.04} & 0.53 & 0.63 & 0.68 & 0.74\\\hdashline
         $\text{MMD-ADiTT}$ &  & \textbf{0.10} & \textbf{0.63} & \textbf{0.85} & \textbf{0.90} & \textbf{0.93}\\
         $\text{MMD-ADiTE}$ & 256 & 0.39 & 0.59 & 0.77 & 0.74 & 0.89\\
         $\text{MMD-Sub}$ &  & 0.16 &0.55 & 0.67 & 0.75 & 0.76\\\hdashline
         $\text{MMD-ADiTT}$ &  & \textbf{0.10} & \textbf{0.69} & \textbf{0.92} & \textbf{0.94} & \textbf{0.97}\\
         $\text{MMD-ADiTE}$ & 512 & 0.27 & 0.58 & 0.77 & 0.71 & 0.85\\
         $\text{MMD-Sub}$ &  & 0.12 &0.62 & 0.76 & 0.83 & 0.86\\\hdashline
         $\text{MMD-ADiTT}$ &  & \textbf{0.12} & \textbf{0.80} & \textbf{0.95} & \textbf{0.97} & \textbf{0.97}\\
         $\text{MMD-ADiTE}$ & 1024 & 0.32 & 0.65 & 0.80 & 0.72 & 0.87\\
         $\text{MMD-Sub}$ &  & \textbf{0.12} &0.68 & 0.89 & 0.94 & 0.95\\\hdashline
         $\text{MMD-ADiTT}$ &  & 0.11 & \textbf{0.91} & \textbf{0.98} & \textbf{0.98} & \textbf{0.98}\\
         $\text{MMD-ADiTE}$ & 2048 & 0.20 & 0.69 & 0.76 & 0.67 & 0.89\\
         $\text{MMD-Sub}$ &  & \textbf{0.08} &0.79 & 0.96 & 0.95 & \textbf{0.98}\\
        \hline
    \end{tabular}
\end{table*}
 % data nulls
 % data alts
% \begin{figure}
%     \centering
%     \includegraphics[width=0.84\linewidth]{figs/exps_1/heatmaps_aditt.png}
%     \caption{.}
%     \label{fig:tv_aditt_heatmap}
% \end{figure} % heatmaps aditt
 % heatmaps sub
% \begin{figure}
%     \begin{subfigure}{0.49\linewidth}
%       \includegraphics[trim={5mm 5mm 3mm 5mm}, clip, width=0.8\linewidth]{figs/exps_1/marginal_ref_weights.pdf}
%       \caption{\small{.}}
%       \label{fig:tv_marginal_ref}
%     \end{subfigure}
%     \begin{subfigure}{0.49\linewidth}
%       \centering
%       \includegraphics[trim={5mm 5mm 3mm 5mm}, clip, width=1.02\linewidth]{figs/exps_1/marginal_test_weights.pdf}
%       \caption{\small{.}}
%       \label{fig:breeds_power}
%     \end{subfigure}
%       \caption{Caption}
%       \label{fig:tv_marginal_test}
%     \end{figure} % marginal weights
 % perms
 % qqs
\begin{figure}
    \centering
\includegraphics[width=0.8\linewidth]{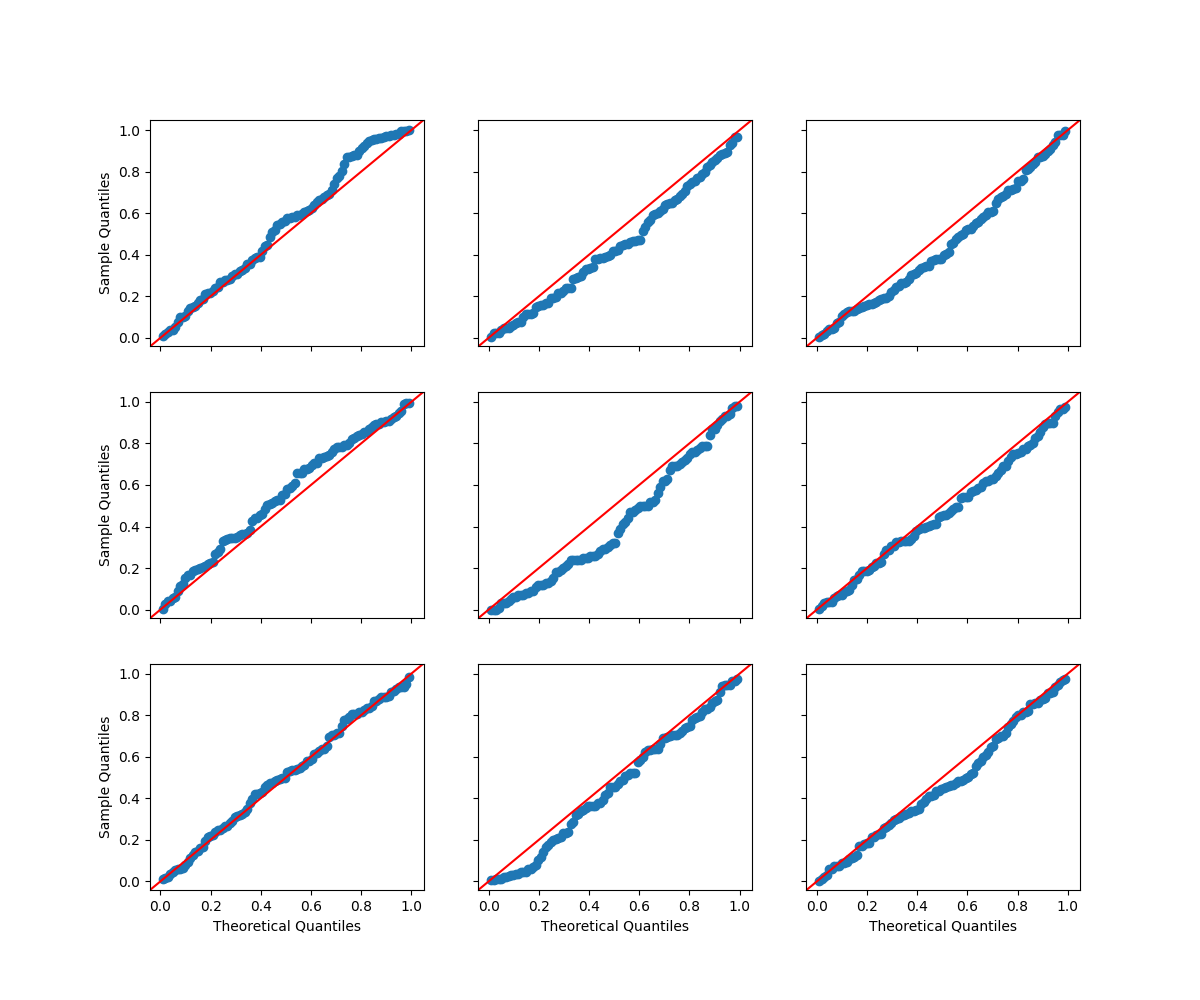}
\caption{Shown centrally is the Q-Q plot of a $U[0,1]$ against the p-values obtained by MMD-Sub under a change in context distribution from $N(0,1)$. The context-conditional distribution has not changed and therefore a perfectly calibrated detector should have a Q-Q plot lying close to the diagonal. To contextualise how well the central plot follows the diagonal, we surround it with Q-Q plots corresponding to 100 p-values actually sampled from $U[0,1]$.}
\label{fig:tv_qqs_sub}
\end{figure} % qqs

\subsection{Controlling for subpopulation prevalences}\label{app:4_2}

\begin{figure}
    \centering
    \includegraphics[width=0.9\linewidth]{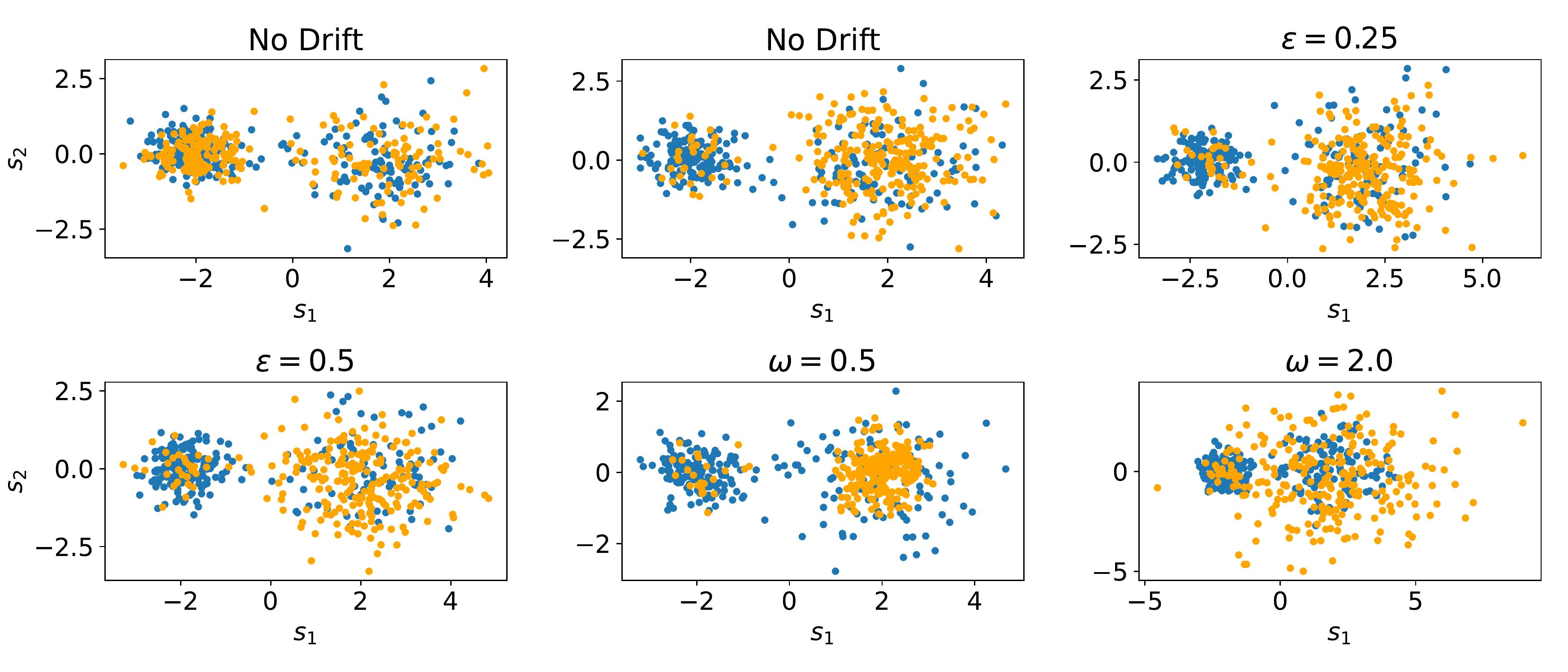}
      \caption{Visualisation of reference (blue) and deployment (orange) instances under various no drift/drift scenarios where we would like to allow the prevalence of modes in a mixture of two Gaussians to vary, but not their underlying distributions. From top left to bottom right: no change to either the prevalence of modes or their underlying distribution; a change only to the prevalence of modes; a change in the prevalence of modes as well as a shift in the mean of one mode by $\epsilon=0.25$; a change in the prevalence of modes as well a shift in the mean of one mode by $\epsilon=0.5$; a change in the prevalence of modes as well a scale in standard deviation for one mode by $\omega=0.5$; a change in the prevalence of modes as well a scale in standard deviation for one mode by $\omega=2.0$.}
      \label{fig:blobs_data}
\end{figure}

For this problem we take $\mathcal{S}=\mathbb{R}^2$ and a (context-unconditional) reference distribution of $P_{S_0}=\pi_1N(\mu_1,\sigma_1^2\mathbf{I}) + \pi_2N(\mu_2,\sigma_2^2\mathbf{I})$ where $\mathbf{I}$ is the 2-dimensional identity matrix. We aim to detect drift in the distributions underlying subpopulations, i.e. changes in $(\mu_1,\mu_2,\sigma_1,\sigma_2)$, whilst allowing their relative prevalences, i.e. $(\pi_1, \pi_2)$, to change. For each run we sample new prevalences $(\pi_1,\pi_2)$ from a $\text{Beta}(2,2)$, means $\mu_1,\mu_2$ from a $N(0,\mathbf{I})$ and variances from an inverse gamma distribution with mean 0.5, such that some runs have highly overlapping modes and others do not. An example, for a run with minimal mode overlap, is shown in Figure~\ref{fig:blobs_data}. 

 We do not assume access to knowledge of the subpopulation from which samples were generated. We therefore first perform unsupervised clustering to associate samples with a vector containing the probability that it belongs to each subpopulation. We do this by fitting, in an unsupervised manner, a Gaussian mixture model (GMM) with 2 components. The GMM is fit using a held out portion of the reference data. We hold out the same amount of reference data -- 25\% -- that is held out of the deployment data by the MMD-ADiTT to condition CoDiTE functions on and by MMD-Sub to fit density estimators. Because we consider only two subpopulations (for illustrative purposes) we can simply use the first element of the vector of two subpopulation probabilities as the context variable $C\in[0,1]$, which we can think of as a proxy for subpopulation membership. 
 
\begin{figure}
    \centering
\includegraphics[width=0.8\linewidth]{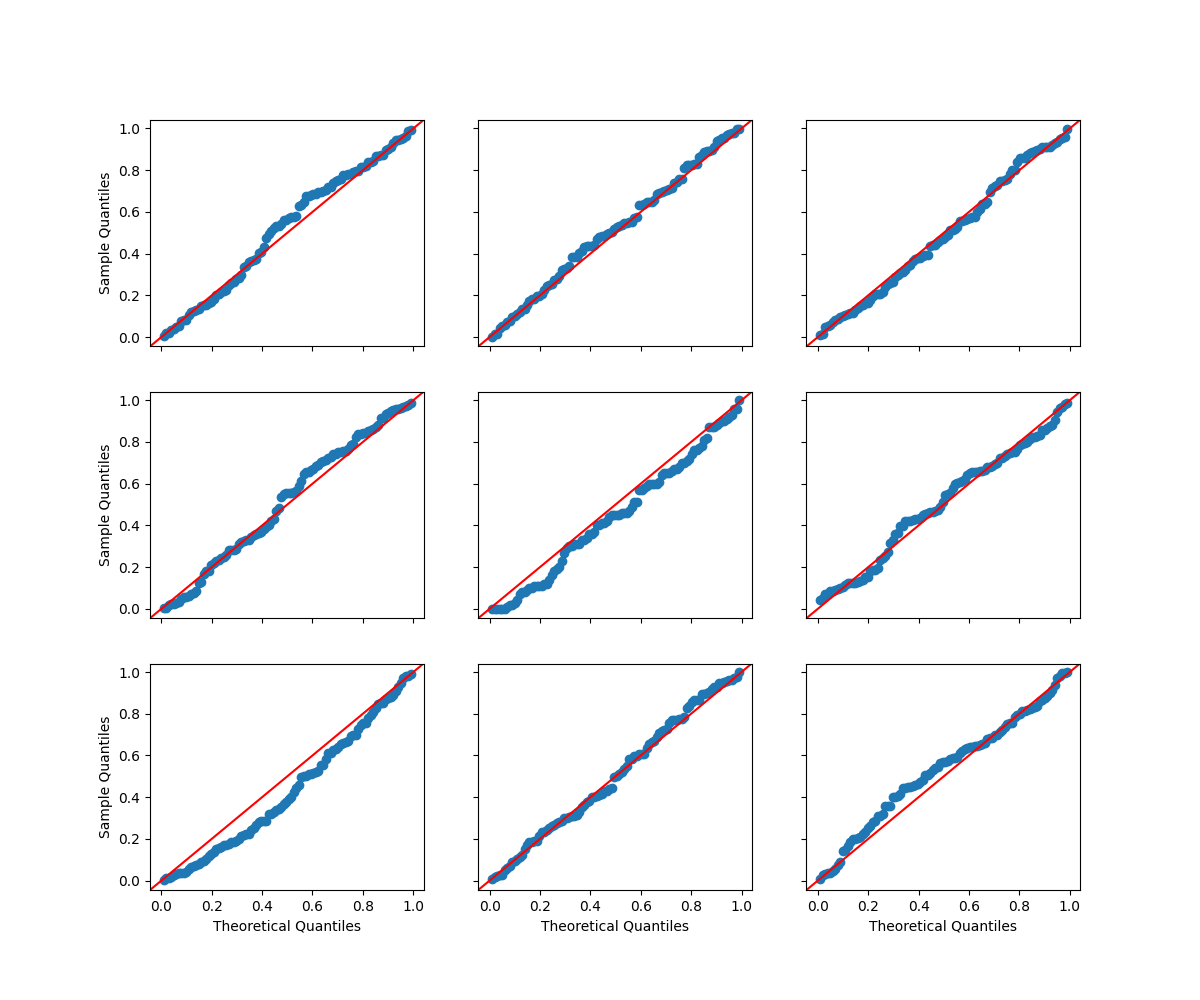}
\caption{Shown centrally is the Q-Q plot of a $U[0,1]$ against the p-values obtained by MMD-ADiTT under a change to the prevalence of modes in a Gaussian mixture. The context-conditional distribution has not changed and therefore a perfectly calibrated detector should have a Q-Q plot lying close to the diagonal. To contextualise how well the central plot follows the diagonal, we surround it with Q-Q plots corresponding to 100 p-values actually sampled from $U[0,1]$.}
\label{fig:blobs_qqs}
\end{figure}

\begin{figure}
    \centering
\includegraphics[width=0.8\linewidth]{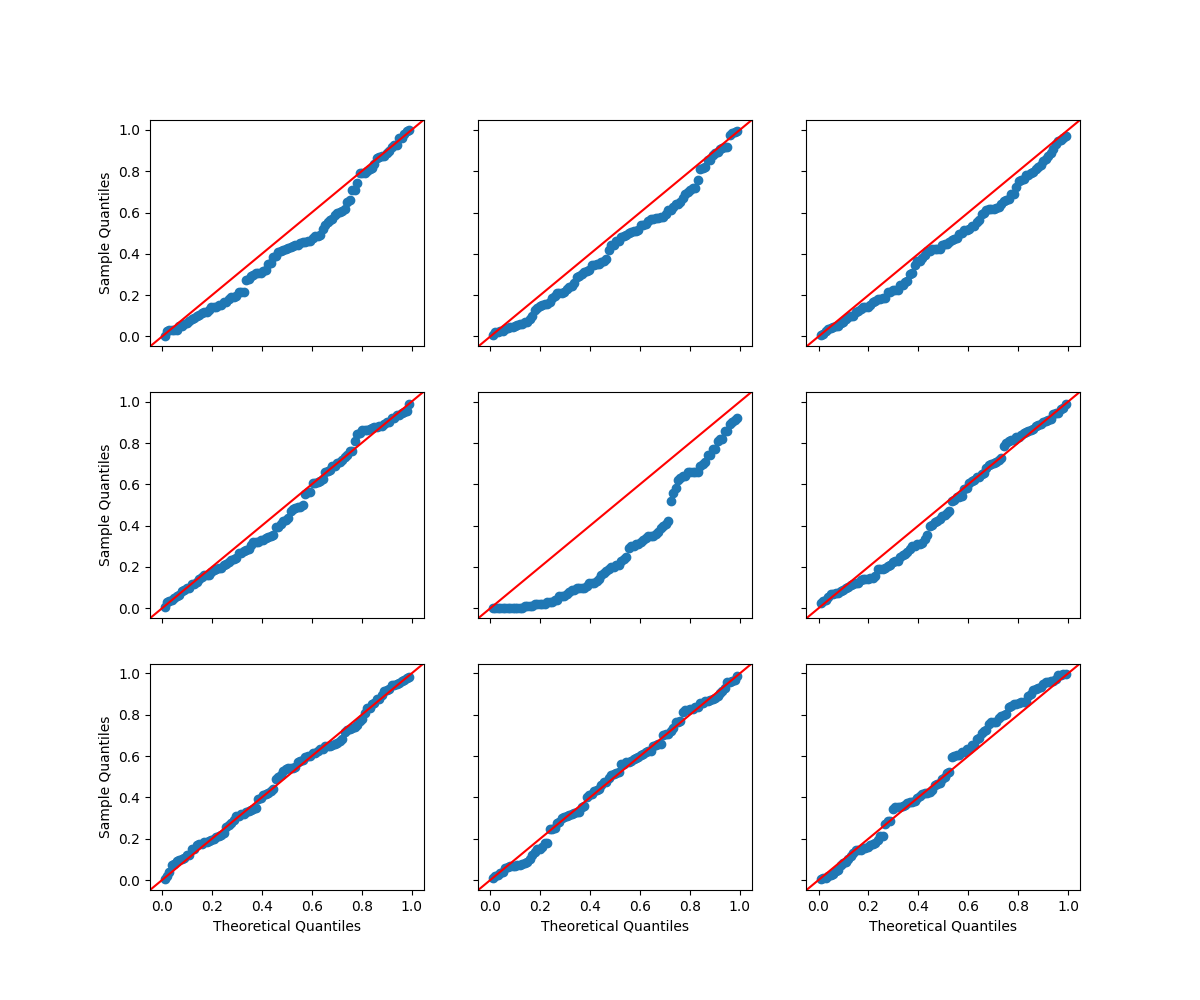}
\caption{Shown centrally is the Q-Q plot of a $U[0,1]$ against the p-values obtained by MMD-Sub under a change to the prevalence of modes in a Gaussian mixture. The context-conditional distribution has not changed and therefore a perfectly calibrated detector should have a Q-Q plot lying close to the diagonal. To contextualise how well the central plot follows the diagonal, we surround it with Q-Q plots corresponding to 100 p-values actually sampled from $U[0,1]$.}
\label{fig:blobs_qqs_sub}
\end{figure}

 To test the calibration of detectors as subpopulation prevalences vary we sample, for each run, new prevalences $(\omega_1,\omega_2)$ from a $\text{Beta}(1,1)$. We choose $\text{Beta}(2,2)$ and $\text{Beta}(1,1)$ to generate reference and deployment prevalences respectively so that deployment distibutions are typically more extremely dominated by one subpopulation than referencee distributions, as would be common in practice. Table~\ref{tab:blobs_results} shows, for various sample sizes, the calibration of the MMD-ADiTT and MMD-Sub detectors as prevalences vary in this manner. We see that MMD-ADiTT achieves strong calibration whereas MMD-Sub does not. Figures~\ref{fig:blobs_qqs} and ~\ref{fig:blobs_qqs_sub} further demonstrate the difference in calibration. 

\begin{figure}
    \centering
    \includegraphics[width=0.75\linewidth]{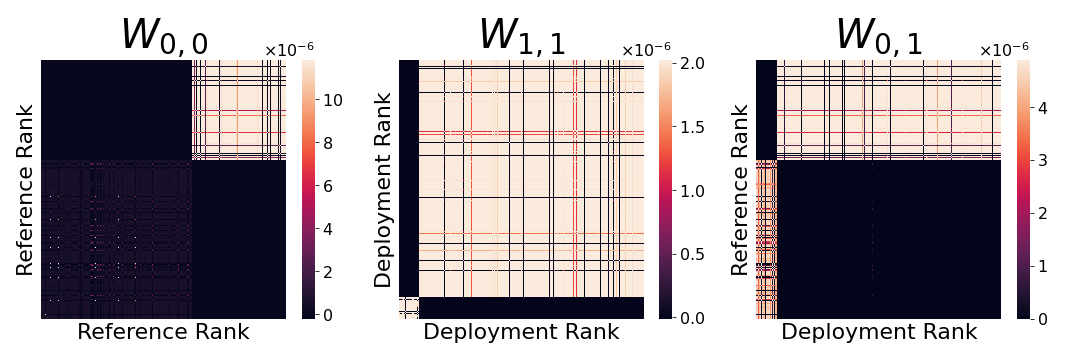}
    \caption{Visualisation of the weight matrices used in computation of the MMD-ADiTT when deployment contexts fall into two disjoint modes. 
    Here we order samples by subpopulation (rather than context) to show explicitly that conditioning on proxies managed to achieve the desired block structure where only similarities between instances in the same same subpopulation contribute. The shapes of the blocks correspond to reference and deployment subpopulation prevalences.}
    \label{fig:blobs_aditt_heatmap}
\end{figure}

\begin{figure}
    \centering
    \includegraphics[width=0.75\linewidth]{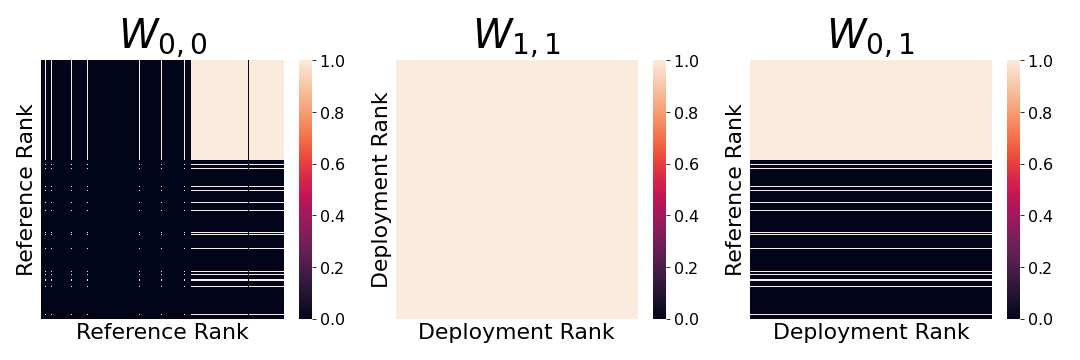}
    \caption{Visualisation of the weight matrices used in computation of the MMD-Sub when deployment contexts fall into two disjoint modes. 
    Here we order samples by subpopulation (rather than context) and again see how similarities between instances in different modes contribute.}
    \label{fig:blobs_sub_heatmap}
\end{figure}

\begin{figure}
\centering
    \begin{subfigure}{0.44\linewidth}
    \centering
      \includegraphics[trim={5mm 5mm 3mm 5mm}, clip, width=1.02\linewidth]{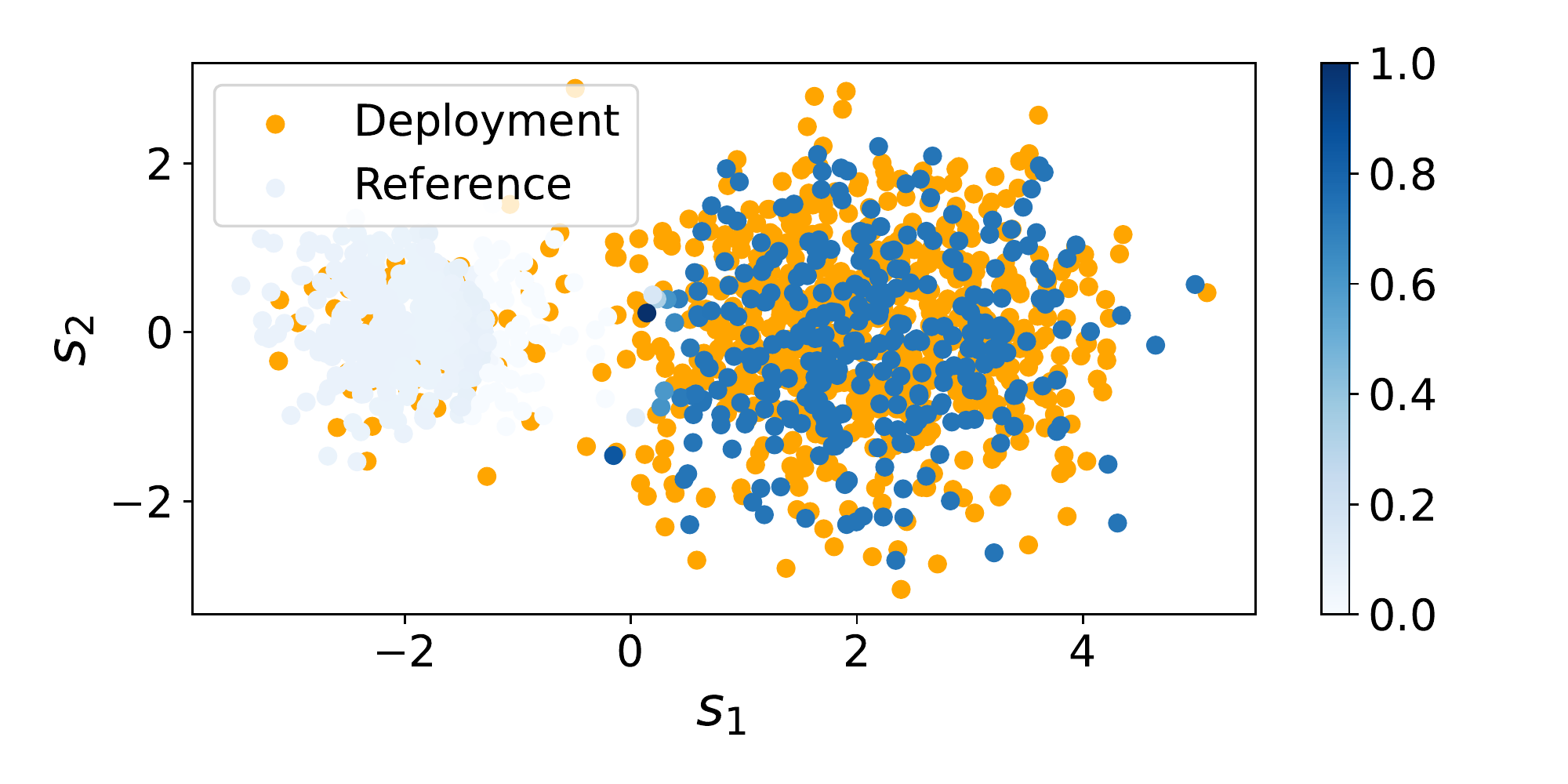}
      \caption{\small{}}
      \label{fig:blobs_marginal_ref}
    \end{subfigure}
    \begin{subfigure}{0.44\linewidth}
      \centering
      \includegraphics[trim={5mm 5mm 3mm 5mm}, clip, width=1.02\linewidth]{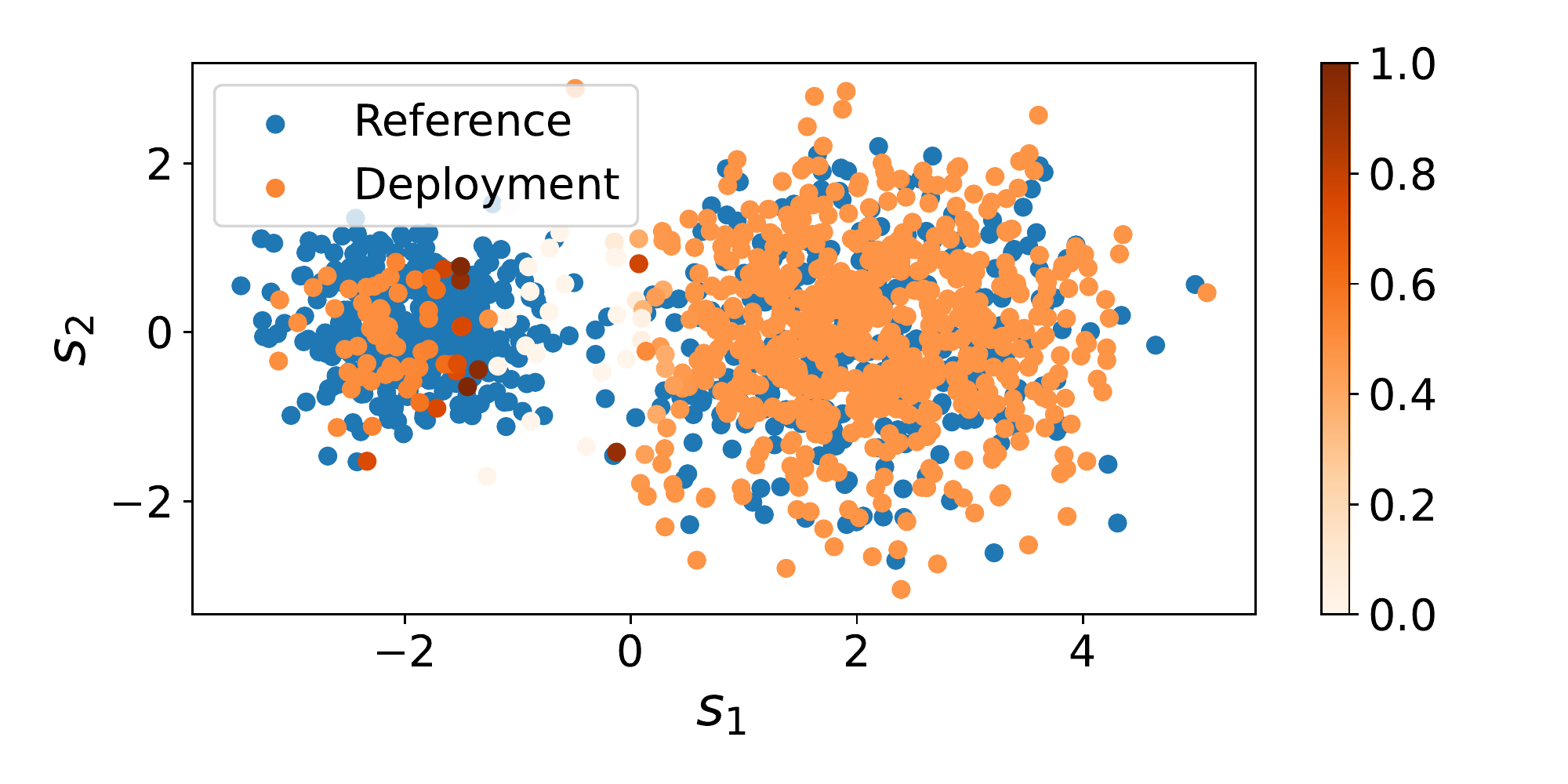}
      \caption{\small{}}
      \label{fig:blobs_marginal_test}
    \end{subfigure}
    \caption{Visualisation of the weight attributed by MMD-ADiTT to comparing each reference sample to the set of deployment samples (left) and vice versa (right). Only reference samples with contexts in the support of the deployment contexts significantly contribute. Weight here refers to the corresponding row/column sum of $W_{0,1}$.}
      \label{fig:blobs_marginal}
\end{figure}

To test the power of detectors in response to a change in location of the distribution underlying a single subpopulation we randomly select one of the two subpopulations and perturb its mean in a random direction by $\epsilon\in\{0.2,0.4,0.6,0.8,1.0\}$ standard deviations. Similarly to test power to detect change in scale we randomly select one of the two subpopulations and multiply its standard deviation by a factor of $\omega\in\{0.5,1.0\}$. Results are shown in Table~\ref{tab:blobls_results}. Having a univariate context variable again allows us to visualise weight martices, as shown in Figure~\ref{fig:blobs_aditt_heatmap} and Figure~\ref{fig:blobs_sub_heatmap}, where we again observe the desired block structure for MMD-ADiTT and not for MMD-Sub. We also plot the marginal weights assigned by MMD-ADITT to reference and deployment instances in Figure~\ref{fig:blobs_marginal}. Here we see deployment weights receiving fairly uniform weights whereas the reference weights in the left mode receive much less weight than those in the right mode. This is due to the changes in prevalences making it necessary to upweight reference instances in the right mode to match the high proportion of deployment instances in that mode and similarly downweight the reference instances in the left mode to match the lower proportion.
 
\begin{table*}
\small
    \centering
    \caption{Calibration under changes to the mixture weights of a mixture of two Gaussians and power under a change corresponding to shifting one of the components by $\epsilon$ standard deviations or scaling its standard deviation by a factor of $\omega$. }\label{tab:blobs_results}
    % \small
    \vspace{2mm}
        \begin{tabular}{| *{10}{c|} }
            \hline
            Method & Sample Size & Calibration (KS) & \multicolumn{7}{c|}{Power (AUC)} \\ 
         \hline
         &  & & $\epsilon=0.2$ & $\epsilon=0.4$ & $\epsilon=0.6$ & $\epsilon=0.8$ & $\epsilon=1.0$ & $\omega=0.5$ & $\omega=2.0$\\
         \hline
         $\text{MMD-ADiTT}$ & 128 & \textbf{0.13} & \textbf{0.53} & \textbf{0.63} & \textbf{0.72} & \textbf{0.79} & \textbf{0.84} & \textbf{0.79} & \textbf{0.83}\\
         $\text{MMD-Sub}$ & 128 & 0.31 & 0.52 & 0.59 & 0.66 & 0.74 & 0.78 & 0.70 & 0.76 \\\hdashline
         $\text{MMD-ADiTT}$ & 256 & \textbf{0.13} & \textbf{0.56} & \textbf{0.70} & \textbf{0.80} & \textbf{0.88} & \textbf{0.91} & \textbf{0.89} & \textbf{0.91}\\
         $\text{MMD-Sub}$ & 256 & 0.28 & 0.55 & 0.65 & 0.75 & 0.81 & 0.84 & 0.81 & 0.83\\\hdashline
         $\text{MMD-ADiTT}$ & 512 & \textbf{0.09} & 0.56 & \textbf{0.75} & \textbf{0.88} & \textbf{0.92} & \textbf{0.93} & \textbf{0.92} & \textbf{0.95}\\
         $\text{MMD-Sub}$ & 512 & 0.35 & \textbf{0.57} & 0.71 & 0.80 & 0.84 & 0.86 & 0.83 & 0.83\\\hdashline
         $\text{MMD-ADiTT}$ & 1024 & \textbf{0.15} & \textbf{0.67} & \textbf{0.84} & \textbf{0.91} & \textbf{0.94} & \textbf{0.96} & \textbf{0.96} & \textbf{0.96}\\
         $\text{MMD-Sub}$ & 1024 & 0.32 & 0.63 & 0.83 & 0.88 & 0.90 & 0.91 & 0.87 & 0.90\\\hdashline
         $\text{MMD-ADiTT}$ & 2048 & \textbf{0.12} & \textbf{0.78} & \textbf{0.91} & \textbf{0.94} & \textbf{0.95} & \textbf{0.96} & \textbf{0.94} & \textbf{0.95}\\
         $\text{MMD-Sub}$ & 2048 & 0.38 & 0.70 & 0.82 & 0.86 & 0.87 & 0.87 & 0.85 & 0.86\\
        \hline
    \end{tabular}
\end{table*}
 % data
 % heatmap aditt
 % heatmap sub
 % marginal weights
\begin{figure}
    \centering
\includegraphics[width=0.8\linewidth]{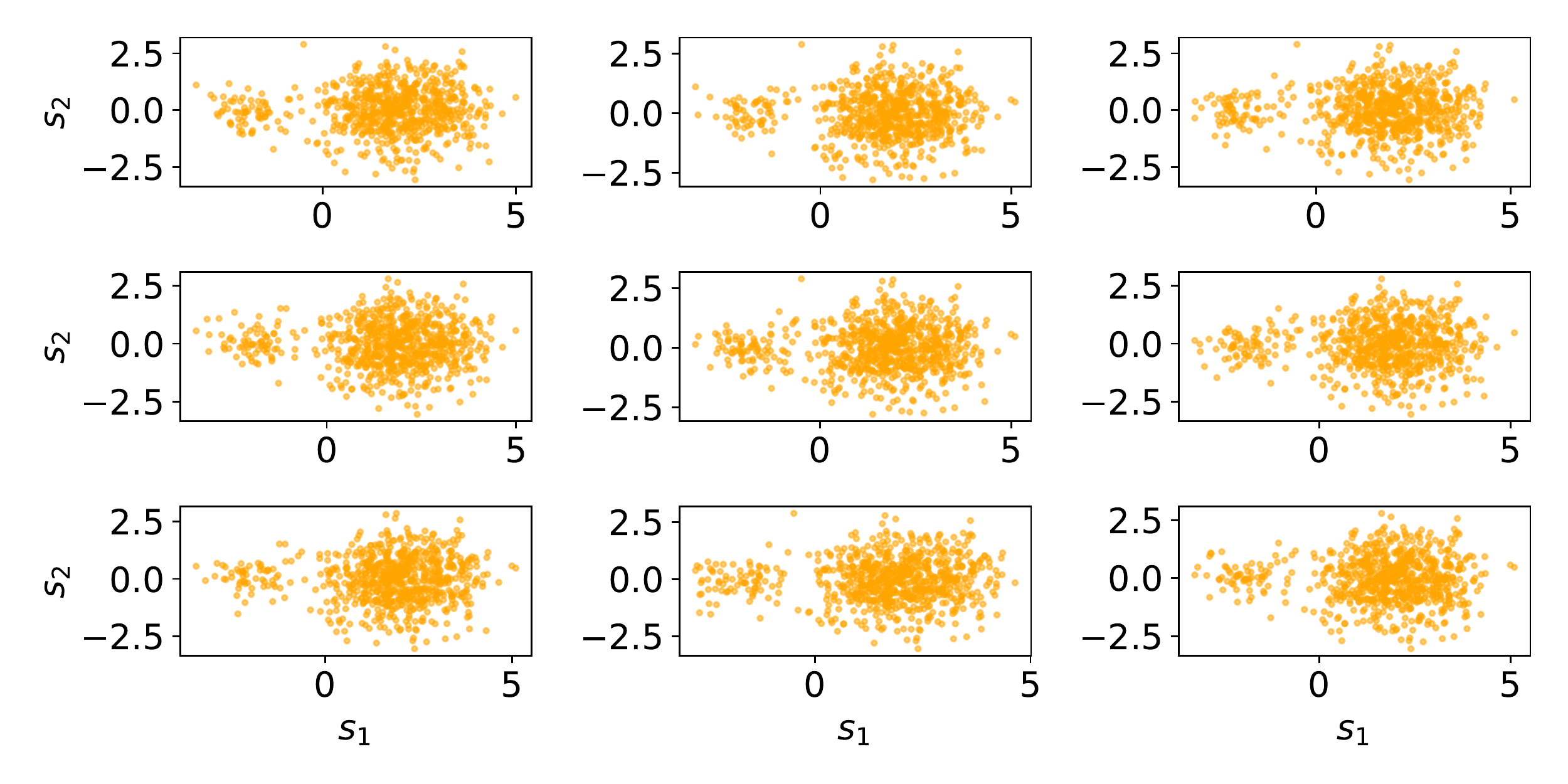}
\caption{The central plot shows here shows a batch of deployment samples shown generated under the same setup as Figure~\ref{fig:blobs_data}. The surrounding plots all show alternative sets of reassigned deployment samples obtained by using the conditional resampling procedure of \citet{rosenbaum1984conditional} to reassign deployment statuses as $z'_i\sim\text{Ber}(e(c_i))$ for $i=1,...,n$. Note that the alternatives do not use identical samples for each reassignment, but do manage to achieve the desired context distribution, with none of the plots being noticeably different from any other.}
\label{fig:blobs_perms}
\end{figure} % perms
 % qqs
 % qqs

\subsection{Controlling for model predictions}\label{app:4_3}

\citet{santurkar2021breeds} organise the 1000 ImageNet \cite{deng2009imagenet} classes into a hierarchy of various levels. At level 2 of their hierarchy exists 10 superclasses each containing a number of semantically related subclasses. We retain the 6 superclasses containing 50 or more subclasses, to allow sufficient samples for our experiments. Anew for each run, for each superclass we sample 25 subclasses to act as the reference distribution of the superclass and 25 subclasses to act a drifted alternative. We use the ImageNet training split to train a model $M$ to predict superclasses from the undrifted samples. We then use the ImageNet validation split for experiments, assigning images $x$ model predictions in $M(x)\in[0,1]^6$ as context $c$. We wish to allow the distribution of model predictions to change between reference and deployment samples so long as the distribution of images, conditional on the model predictions, remains the same.

The model is defined as $M(x)=H(B(x))$, where $B$ is the convolutional base of a pretrained SimCLR \cite{chen2020simple} model, which maps images onto 2048-dimensional vectors, and $H(x):\mathbb{R}^{2048}\rightarrow[0,1]^6$ is a classification head. We define $H(x)=\text{Softmax}(L_2(\sigma(L_1(x))))$ to consist of a linear projection onto $\mathbb{R}^{128}$, followed by a ReLU activation, followed by a linear projection onto $\mathbb{R}^6$, followed by a softmax activation. Each of the 100 runs uses different subclasses for superclasses and therefore we retrain $H$ on the full ImageNet training set for each run. We do so for just a single epoch, which was invariably sufficient to obtain a classifier with an accuracy of between 91\% and 93\%. 

We take the reference context distribution to be that corresponding to the model's predictions across all images. To test calibration we vary this for the deployment context distribution by selecting $K$ of 6 superclasses and only retaining contexts for which the superclass deemed most probable is one of the $K$ chosen. Therefore when $K=1$, the context distribution has collapsed onto around one sixth of its original support, whereas for $K=6$ it has not changed. Table~\ref{tab:breeds_calib} shows, for a sample size of $n_0=n_1=1000$, how well calibrated detectors are under such changes.

To test power to detect changes in the distribution of images conditional on model predictions we keep the deployment context distribution the same as the reference context distribution at $K=6$ so that conventional MMD two-sample tests can be compared to. We then change the distribution underlying $J$ of the 6 superclasses to their drifted alternatives. Table ~\ref{tab:blobls_results} shows how power varies with $J$.

\begin{table*}
\small
    \centering
    \caption{Calibration under the change of context corresponding to a computer vision model making predictions evenly across 6 classes to making predictions only within $K$ of 6 classes. }\label{tab:breeds_calib}
    \vspace{2mm}
    % \small
        \begin{tabular}{| *{10}{c|} }
            \hline
            Method  & \multicolumn{6}{c|}{Calibration (KS)} \\ 
         \hline
          & $K=1$ & $K=2$ & $K=3$ & $K=4$ & $K=5$ & $K=6$\\
         \hline
         $\text{MMD-ADiTT}$  & \textbf{0.08} & \textbf{0.08} & \textbf{0.11} & \textbf{0.14} & \textbf{0.09} & \textbf{0.07}\\
         $\text{MMD-Sub}$ & 0.89 & 0.88 & 0.80 & 0.64 & 0.43 & 0.23  \\
        \hline
    \end{tabular}
\end{table*}
\begin{table*}
\small
    \centering
    \caption{Power to detect changes in the distributions underlying $J$ of the 6 classes being predicted by a computer vision model.}\label{tab:blobls_results}
    \vspace{2mm}
        \begin{tabular}{| *{10}{c|} }
            \hline
            Method & Sample Size & Calibration (KS) & \multicolumn{6}{c|}{Power (AUC)} \\ 
         \hline
         &  & $J=0$ & $J=1$ & $J=2$ & $J=3$ & $J=4$ & $J=5$ & $J=6$\\
         \hline
         $\text{MMD-ADiTT}$ & 128 & 0.19 & 0.55 & \textbf{0.59} & \textbf{0.70} & 0.73 & 0.82 & 0.84\\
         $\text{MMD-Sub}$ & 128 & 0.15 & 0.50 & 0.52 & 0.57 & 0.65 & \textbf{0.84} & 0.72  \\
         $\text{MMD}$ & 128 & \textbf{0.13} & \textbf{0.57} & 0.58 & 0.62 & \textbf{0.74} & 0.79 & \textbf{0.89}\\\hdashline
         $\text{MMD-ADiTT}$ & 256 & \textbf{0.13} & \textbf{0.65} & \textbf{0.83} & \textbf{0.86} & \textbf{0.92} & \textbf{0.95} & \textbf{0.98}\\
         $\text{MMD-Sub}$ & 256 & 0.17 &0.60 & 0.55 & 0.72 & 0.78 & 0.88 & 0.86 \\
         $\text{MMD}$ & 256 & 0.19 &0.59 & 0.65 & 0.78 & 0.87 & 0.94 & \textbf{0.98} \\\hdashline
         $\text{MMD-ADiTT}$ & 512 & \textbf{0.10} & \textbf{0.82} & \textbf{0.94} & \textbf{0.99} & \textbf{1.00} & 0.99 & 0.99\\
         $\text{MMD-Sub}$ & 512 & 0.14 &0.54 & 0.65 & 0.82 & 0.91 & 0.88 & 0.96 \\
         $\text{MMD}$ & 512 & 0.18 &0.58 & 0.73 & 0.92 & 0.96 & \textbf{1.00 }& \textbf{1.00} \\\hdashline
         $\text{MMD-ADiTT}$ & 1024 & \textbf{0.06} & \textbf{0.95} & \textbf{1.00} & \textbf{1.00} & \textbf{1.00} & \textbf{1.00} & \textbf{1.00}\\
         $\text{MMD-Sub}$ & 1024 & 0.20 &0.62 & 0.77 & 0.87 & 0.92 & 0.88 & 0.94 \\
         $\text{MMD}$ & 1024 & 0.15 &0.65 & 0.88 & 0.98 & \textbf{1.00} & \textbf{1.00} & \textbf{1.00} \\\hdashline
         $\text{MMD-ADiTT}$ & 2048 & 0.10 & \textbf{0.97} & \textbf{0.98} & 0.98 & 0.98 & 0.98 & 0.98\\
         $\text{MMD-Sub}$ & 2048 & \textbf{0.09} &0.79 & 0.88 & 0.94 & 0.95 & 0.85 & 0.96 \\
         $\text{MMD}$ & 2048 & 0.12 &0.69 & \textbf{0.98} & \textbf{1.00} & \textbf{1.00} & \textbf{1.00} & \textbf{1.00} \\
        \hline
    \end{tabular}
\end{table*}

%%%%%%%%%%%%%%%%%%%%%%%%%%%%%%%%%%%%%%%%%%%%%%%%%%%%%%%%%%%%%%%%%%%%%%%%%%%%%%%
%%%%%%%%%%%%%%%%%%%%%%%%%%%%%%%%%%%%%%%%%%%%%%%%%%%%%%%%%%%%%%%%%%%%%%%%%%%%%%%

\end{document}